\documentclass[10pt,twocolumn,letterpaper]{article}

\usepackage[pagenumbers]{iccv} 

\usepackage{multirow}
\usepackage[T1]{fontenc}
\usepackage{multicol}
\usepackage{makecell}
\usepackage{pifont}
\def\cmark{\ding{51}} 
\def\xmark{\ding{55}}

\newcommand{\nap}{\color{gray}N/A}%
\newcommand{\lna}{{\nap}\hphantom{.}}
\newcommand{\rna}{{\nap}}

\usepackage{tikz}
\usepackage{graphicx}

\usepackage{pgfplots}
\usepgfplotslibrary{colorbrewer}
\pgfplotsset{cycle list/Set1-9}
\tikzset{every picture/.style={line width=1pt}}
\usetikzlibrary{patterns}
\usepgfplotslibrary{fillbetween}
\usepackage{color,soul}

\pgfplotsset{compat=1.18}
\pgfplotsset{
tick label style = {font=\sffamily\scriptsize},
every axis label = {font=\sffamily\scriptsize},
legend style = {font=\sffamily\scriptsize},
}

\usetikzlibrary{shadings} 
\usetikzlibrary[shadings] 

\definecolor{iccvblue}{rgb}{0.21,0.49,0.74}
\usepackage[pagebackref,breaklinks,colorlinks,allcolors=iccvblue]{hyperref}

\title{SMARTIES: Spectrum-Aware Multi-Sensor Auto-Encoder \\for Remote Sensing Images}
\author{Gencer Sumbul \qquad Chang Xu \qquad Emanuele Dalsasso \qquad Devis Tuia \\
Ecole Polytechnique Fédérale de Lausanne (EPFL), Switzerland
}

\begin{document}
\maketitle
\begin{abstract}
From optical sensors to microwave radars, leveraging the complementary strengths of remote sensing (RS) sensors is crucial for achieving dense spatio-temporal monitoring of our planet. In contrast, recent deep learning models, whether task-specific or foundational, are often specific to single sensors or to fixed combinations: adapting such models to different sensory inputs requires both architectural changes and re-training, limiting scalability and generalization across multiple RS sensors. On the contrary, a single model able to modulate its feature representations to accept diverse sensors as input would pave the way to agile and flexible multi-sensor RS data processing. To address this, we introduce SMARTIES, a generic and versatile foundation model lifting sensor-specific/dependent efforts and enabling scalability and generalization to diverse RS sensors: SMARTIES projects data from heterogeneous sensors into a shared spectrum-aware space, enabling the use of arbitrary combinations of bands both for training and inference. To obtain sensor-agnostic representations, we train a single, unified transformer model reconstructing masked multi-sensor data with cross-sensor token mixup. On both single- and multi-modal tasks across diverse sensors, SMARTIES outperforms previous models that rely on sensor-specific pretraining. Our code and pretrained models are available at \url{https://gsumbul.github.io/SMARTIES}.
\end{abstract}

\section{Introduction}
\label{sec:intro}
\begin{figure}[t]
  \centering
  \begin{subfigure}{\linewidth}
\includegraphics[width=\linewidth]{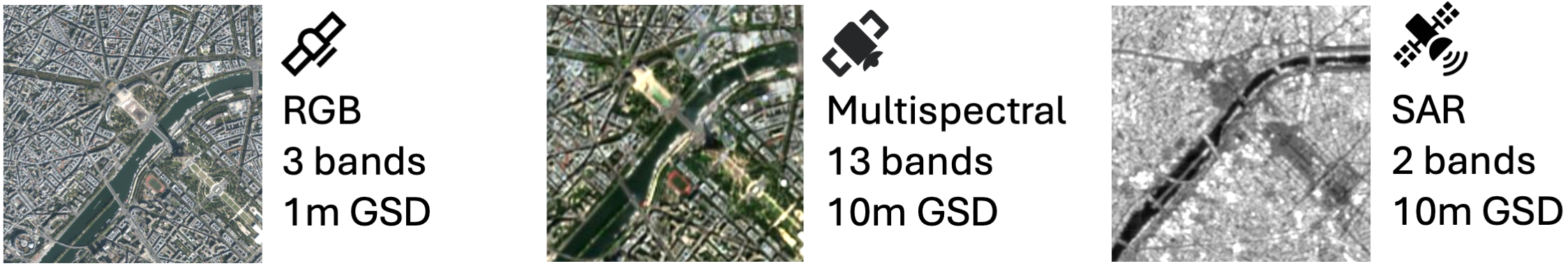}
    \caption{}
    \label{fig:short-a}
  \end{subfigure}
  \vfill
  \begin{subfigure}{\linewidth}
\includegraphics[width=\linewidth]{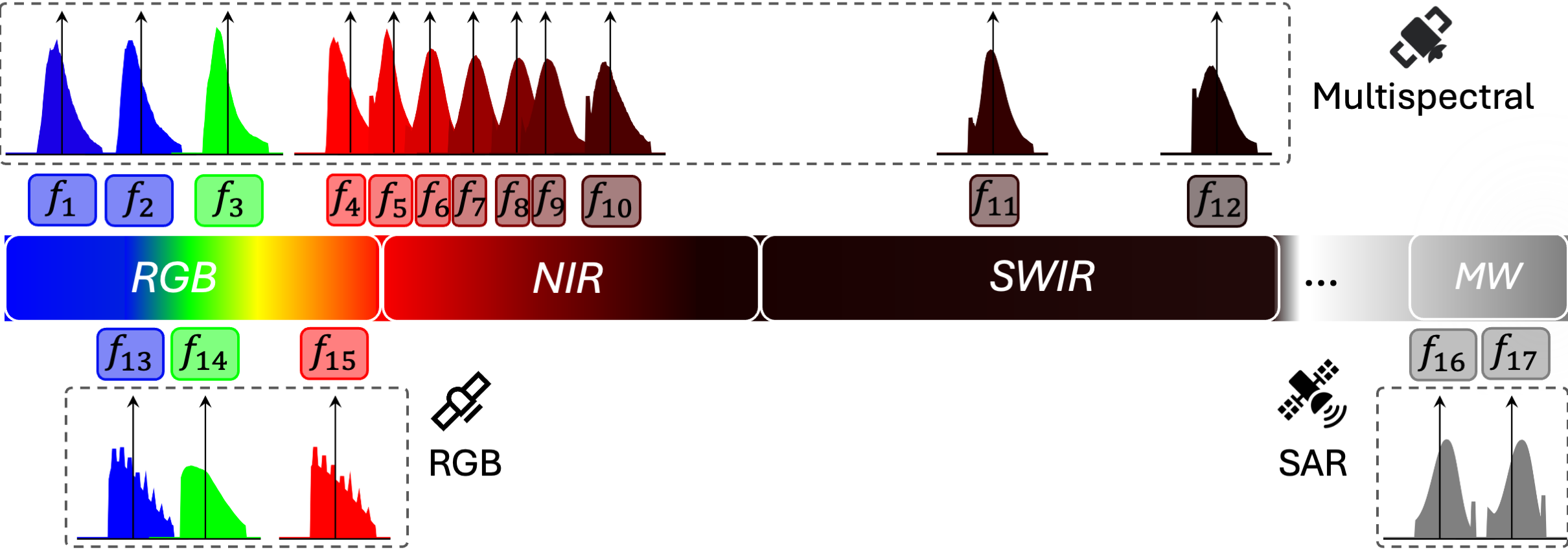}
    \caption{}
    \label{fig:short-b}
  \end{subfigure}
  \caption{(a) An example of RGB, multispectral, and SAR images representative of the different spectral and spatial properties of RS sensors. (b) The spectral bands' histograms for each sensor are shown as probability density function estimations, aligned with the corresponding wavelength range in the electromagnetic spectrum (shown in log scale). SMARTIES leverages different projection layers $\{f_1,f_2,\ldots,f_{17}\}$ for different spectral ranges that allow a single, unified model independent from sensors specificities.}
  \label{fig:short}
\end{figure}

Every day, a vast number of airborne and spaceborne sensors generate tens of terabytes of remote sensing (RS) data, empowering Earth observation~\cite{rsbigdata_ieee} (\cref{fig:short}a). 
Unlike RGB natural images, RS data spans a wide spectrum of wavelengths captured by diverse sensors from RGB visible light, through infrared frequencies and all the way to Microwaves (\cref{fig:short}b).
RS sensors capture electromagnetic radiation at different frequencies: optical sensors capture (very) high-resolution images mostly in the visible and infrared spectrum, offering rich semantic details, but limited by clouds and daylight; synthetic aperture radars (SAR) are active sensors that can acquire images day and night and independently of weather conditions; thermal sensors measure the energy emitted directly by objects at the surface and can be used to estimate surface temperature.
The complementary strengths of these 
sensors hold the potential for continuous, all-day, and all-weather Earth observation, supporting a wide range of applications in various fields, \textit{e.g.}, agriculture, climate, hydrology, urban planning~\cite{multimodalrs_survey_ieee,Tui20grsm}.
Despite the wealth of RS sensors, a significant barrier remains: the lack of unified image representations for multi-modal processing of RS data.
A significant obstacle to achieve this goal originates from the highly heterogeneous characteristics of RS sensors, in terms of spectral range, radiometric resolution, and spatial resolution: such diversity forced previous attempts to design restrictive sensor-specific models~\cite{satmae,spectral_gpt,croma,s2mae,satmae_pp}.
To mitigate this issue, several foundation models (FMs) have been proposed by designing sensor-specific backbones, leading to an increase in computational complexity~\cite{croma, dofa_2024_arxiv,skysense}. Adding new sensors at pretraining and finetuning would require modifying the architecture with extra backbones, leading to further computational overhead, and thus limiting the \textit{scalability} of such models. Moreover, models trained on a fixed combination of sensors will develop biases towards them, suffering from limited \textit{generalization} to unseen sensors.

To address these challenges, we propose a sensor-agnostic FM named \textbf{S}pectrum-Aware \textbf{M}ulti-Sensor \textbf{A}uto-Encoder for \textbf{R}emo\textbf{t}e Sensing \textbf{I}mag\textbf{es} (\textbf{SMARTIES}) that breaks the representation barriers between sensors\footnote{We focus only on amplitude-related phenomenology for RS sensors excluding, for instance, phase information recorded by SAR instruments.} and enables downstream applications using a single, unified model across diverse sensors, including unseen sensor transfer capabilities.
To train a single model on heterogeneous sensors efficiently, we unify sensor representations by projecting data into a shared and divisible space called the spectrum-aware space. 
This concept is based on the observation that, despite the varying spectral ranges, all the different sensors capture subsets of the full electromagnetic spectrum with well-defined physical properties: an example of three typical RS sensors and of the spectral ranges of their bands is shown in Fig.~\ref{fig:short}. Given a specific sensor, each one of its bands is projected into the spectrum-aware space by projection layers specific to wavelength ranges ($f$ in \cref{fig:short}).
By representing data with this shared space, we eliminate the need to train a separate model for each sensor. Moreover, SMARTIES can generalize to unseen sensors during inference by interpolating the learned projection layers to represent the unseen wavelength ranges when full finetuning is costly to achieve. Leveraging this unified representations, we pretrain a transformer model with a self-supervised objective: we learn with cross-sensor token mixup on paired multi-sensor data and enforce the model to reconstruct randomly masked regions of the representations in the spectrum-aware space. Learning this way, SMARTIES maximizes synergies among diverse sensory inputs, scaling efficiently to large training datasets with multiple sensors and enhancing more generalizable representations.

We perform experiments on ten datasets composed of various combinations of sensors, all using the same pretrained backbone. Results demonstrate the superior performance of SMARTIES on both single- and multi-modal tasks across diverse sensors, as well as generalization to new unseen sensors during inference. SMARTIES contributes significantly to literature since: (1) SMARTIES is a single, unified FM without sensor-specific pretraining or backbones that can seamlessly tackle diverse sensory inputs (both single- and multi-modal). (2) SMARTIES exhibits \textit{scalability} to diverse sensors with a high pretraining efficiency. 
Using a ViT encoder, SMARTIES is pretrained with only 500K multispectral, SAR and submeter RGB images and with as little as 300 epochs, showing a smaller computational cost than most RS FMs. 
(3) SMARTIES transfers not only to different downstream applications, but also to new sensors that were not present during pretraining, demonstrating unprecedented \textit{generalization} capabilities.

\section{Related Work}
\label{sec:formatting}
\vspace{.1cm}\noindent\textbf{Self-supervised learning (SSL) in RS} has been widely used to learn general data representations that can be transferred to various downstream tasks \cite{Len20,wang2022self,scheibenreif2022contrastive, Sumbul_2022}. It allows to reduce the demand of task-specific models and of labeled data, which are often scarce in RS. Among different SSL approaches, masked image modelling through masked autoencoders (MAEs) has recently received increasing attention due to its ability to scale to larger models together with the increasing amount of training data~\cite{mae_2022_cvpr, Li:2022}.

\vspace{.1cm}\noindent\textbf{Single-modal MAEs} have demonstrated to learn rich task-transferable representations through a SSL strategy: reconstructing masked parts of images. Several MAE-based FMs have been proposed in RS: SatMAE~\cite{satmae} embeds temporal and spectral information as positional encodings during reconstruction, while SpectralGPT~\cite{spectral_gpt} and S2MAE~\cite{s2mae} model the spatial-spectral data as 3D cubes and enforce the masked reconstruction in the 3D space. Scale-MAE~\cite{scale_mae} encodes different spatial scales in positional encodings to learn robust representations across resolutions.
Despite the task-agnostic strengths, these models trained on a specific image modality (e.g., multispectral data) struggle to handle data in others (e.g., SAR data), hampering their usability and flexibility for multi-modal processing of RS data. 

\begin{figure*}[t]
  \centering
   \includegraphics[width=0.99\linewidth]{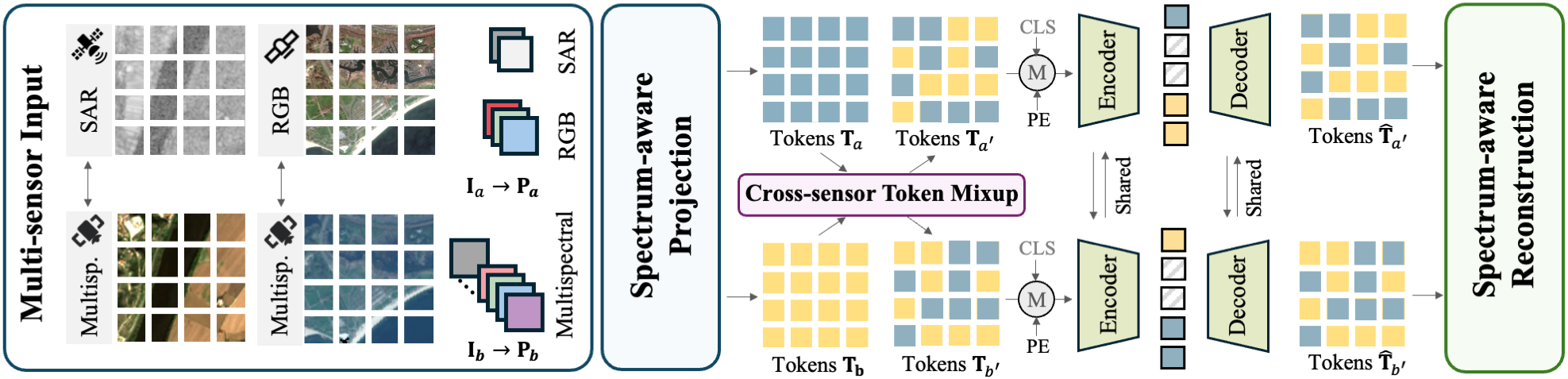}
   \caption{SMARTIES lifts sensor-dependent efforts for multi-sensor RS image representation learning by leveraging: (1) spectrum-aware RS image projection; (2) cross-sensor token mixup; and (3) spectrum-aware RS image reconstruction. PE and Multisp. denote positional encoding and multispectral, respectively.}
   \label{fig:pipeline}
\end{figure*}
\vspace{.1cm}\noindent\textbf{Multi-modal MAEs} extend the MAE framework to develop generalizable representations across different modalities. For example, recent computer vision research focuses on scaling MAE models to accommodate as many image modalities as possible by reconstructing masked multi-modal tokens~\cite{multimae_2022_eccv,4m_nips_2023}. In RS, multi-modal FMs also benefit from MAE-based training~\cite{croma, skysense}. We can identify two groups of models: first, models that learn dual (e.g., CROMA~\cite{croma}) or triple (e.g., SkySense~\cite{skysense}) modal representations with aligned optical and radar sensors; second, models that learn shared features across multiple modalities by blending them into massive training data~\cite{dofa_2024_arxiv,mmearth_eccv_2024}. The first group of models mostly rely on predefined architectural designs for a set of sensors with sensor-dependent encoders, leading to an increase in computational complexity, and limited generalization towards diverse sensors at pretraining and downstream transfer. Although the second group of models can significantly improve generalization across diverse sensors, they need computationally demanding and complex adjustments to MAEs (e.g., hypernetwork for dynamic weight generation in DOFA~\cite{dofa_2024_arxiv}, sensor encodings and channel-specific tokens in SenPa-MAE~\cite{senpa_mae}). In addition, they require massive pretraining sets for learning representations across diverse sensors (8M images for DOFA), showing lower data efficiency, and thus limited scalability.

To fill these gaps, we propose a unified and versatile model that can seamlessly handle diverse sensors during pretraining, exhibiting scalability to diverse sensors. Our model can not only transfer to different downstream tasks on sensors seen during pretraining, but also generalize to new unseen sensors when full finetuning is too expensive.

\section{SMARTIES}
\begin{figure}[t]
  \centering
   \includegraphics[width=\linewidth]{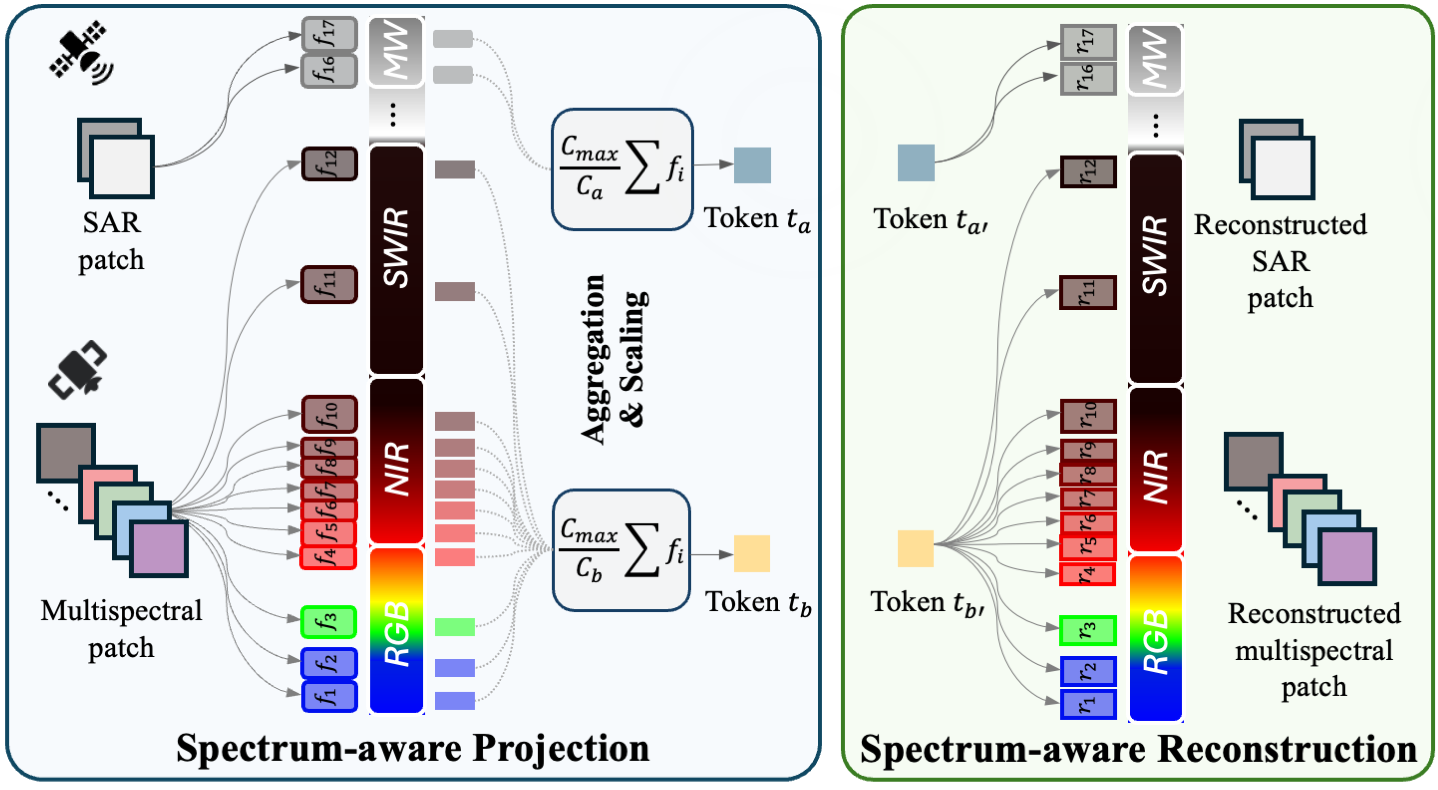}
   \caption{Spectrum-aware RS image projection and reconstruction illustrated on a pair of SAR and multispectral patches.}
   \label{fig:pipeline_spectrum}
\end{figure}

To leverage the multi-sensor nature of RS, we develop a FM named Spectrum-Aware Multi-Sensor Auto-Encoder for Remote Sensing Images (SMARTIES) that learns image representations transferable to diverse sensors. SMARTIES is designed in a \textit{generic} way (\cref{fig:pipeline}), so that it can accommodate variations in sensor characteristics, \textit{data-efficient} for \textit{scalable} pretraining, and also conceptually \textit{simple} for downstream applications. We realize these properties via the following design decisions:
\begin{enumerate}
    \item \textit{Spectrum-aware RS Image Projection} (\cref{saip}): To deal with the variation of spectra in different sensors, we learn spectrum-aware projection layers to tokenize the RS images. These layers depend on the wavelengths of the bands, spanning the continuum of the electromagnetic spectrum covered by RS sensors (\cref{fig:short} and \cref{fig:pipeline_spectrum}).  
    \item \textit{Cross-sensor Token Mixup} (\cref{sec:mixup}): To mitigate the bias specific to sensors or spectrum combinations, we use pairs of aligned images from different sensors as input, and then exchange their tokens. This leads to more scalable and generalizable models.
    \item \textit{Spectrum-aware RS Image Reconstruction} (\cref{sec:rec}): We feed the cross-sensor mixed embeddings into a standard encoder-decoder based transformer, which is easy to deploy for downstream applications. We reproject the decoded images back to the original spectral channels of the RS sensors through spectrum-aware reprojection layers (\cref{fig:pipeline_spectrum}). Spatial and spectral reasoning is employed through masked image modelling with SSL. 
    \item \textit{Downstream Transfer to Diverse Sensors} (\cref{sec:downstream}): Thanks to the spectrum-aware image projection and reconstruction, the resulting encoder can generalize to diverse sensors by using either the existing projection layers or adapting them for unseen sensors by interpolation. 
\end{enumerate}

\subsection{Spectrum-aware RS Image Projection}\label{saip}
SMARTIES uses different projection layers for different spectral ranges, each one defined by the minimum and maximum wavelengths covered by the corresponding band. These ranges can cover various parts of the electromagnetic spectrum (e.g., visible light, near-infrared, short wave infrared, microwaves etc.). This strategy makes each projection layer learn an embedding specific to a certain spectral range, enforcing physical consistency among embeddings of different sensors covering similar spectral ranges. 

By following the official instrument specifications of widely used RS sensors, we define a set of spectrum-aware projection layers $\mathcal{F}=\{f_1, f_2, ..., f_n\}$, where $f_{i}$ is the mapping of the $i$th spectral range via a fully-connected layer. Specifically, we associate the spectral range of each band used in pretraining with a projection layer $f_{i}$: $f_{1}$-$f_{12}$ for Sentinel-2 (S2), $f_{13}$-$f_{15}$ for RGB images of Maxar, and $f_{16}$-$f_{17}$ for Sentinel-1. For instance, $f_{2}$ corresponds to the wavelength range 427nm-558nm of Sentinel-2\footnote{The detailed ranges of projection layers are provided in the \cref{supp:projection_sect} of the supplementary material.}. For bands from different sensors capturing the same band, for instance ``red'' light, but with different frequency ranges (e.g., S2 vs. Maxar), we keep separate projection layers. For a given RS image $\mathbf{I}_{a} \in \mathbb{R}^{W_{a}\times H_{a}\times C_{a}}$ ($C_a$ is the number of spectral bands), we first resize it into the input size $W \times H$ of the model, and then divide it into a sequence of $N_P$ non-overlapping patches $\mathbf{P}_{a} \in \mathbb{R}^{N_W\times N_H\times S^2 C_{a}}$, where $S$ is the patch size, $N_W=W/S$, $N_H=H/S$ and $N_P=N_WN_H$. Each patch $p_a$ of the image $\mathbf{I}_{a}$ is then projected to the joint space using the projection layers $\{f_{i}\}$ in $\mathcal{F}$ corresponding to its bands, where $f_{i}: \mathbb{R}^{S\times S} \rightarrow \mathbb{R}^{D}$ ($D$ is the size of spectrum-aware embeddings). This leads to $C_a$ embeddings (one per band). For token $t_a$ of this patch, all $C_a$ embeddings are first averaged, and then scaled up by $C_{\text{max}}=12$, which is the largest number of spectral bands encountered during pretraining. This last operation prevents imbalance between different sensors due to the different number of bands. The whole process is illustrated in the left panel of \cref{fig:pipeline_spectrum}. We note that thanks to this projection strategy, adding new RS sensors to pretraining simply requires adding new $f_i$s for the additional spectral ranges.

\subsection{Cross-sensor Token Mixup}\label{sec:mixup}
Thanks to the spectrum-aware RS image projection, SMARTIES can operate on RS images acquired by different sensors with a unified approach. However, this would lead to bias towards specific sensors or bands combinations that are more present during pretraining. To alleviate this, we perform cross-sensor token mixup: (1) we first take as input a pair of images acquired by different sensors on the same area, and then (2) we exchange tokens across the images of a pair using mixup, as shown in \cref{fig:pipeline}. This prevents encoding bias towards specific spectral combinations and also enhances the generalization capability of SMARTIES over multiple sensors.
For a multi-sensor image pair $(\mathbf{I}_{a}, \mathbf{I}_{b})$, with tokens $\mathbf{T}_{a}\in \mathbb{R}^{N_W\times N_H\times D}$ and $\mathbf{T}_{b}\in \mathbb{R}^{N_W\times N_H\times D}$, we define the mixed image tokens $\mathbf{T}_{a'}$ as: 
\begin{equation}
\label{eq.mixupa}
    \mathbf{T}_{a'} = \mathcal{M} \odot \mathbf{T}_a + (1-\mathcal{M}) \odot \mathbf{T}_b
\end{equation}
where $\mathcal{M}\in \mathbb{R}^{N_W\times N_H\times D}$ is a randomly generated binary mask broadcasted along the third dimension. We also perform a mirrored mixup to obtain the mirrored version $\mathbf{T}_{b'}$ of $\mathbf{T}_{a'}$ to avoid losing information during the mixup process: 
\begin{equation}
    \mathbf{T}_{b'} = (1-\mathcal{M}) \odot \mathbf{T}_a + \mathcal{M} \odot \mathbf{T}_b.
\end{equation}

\subsection{Spectrum-aware RS Image Reconstruction}\label{sec:rec}
For spatial and spectral reasoning, we use a standard encoder-decoder transformer architecture. Given the lack of labels and in order to remain scalable, we employ self-supervised masked image modelling with spectrum-aware reconstruction. As in the vanilla MAE, we follow the \textbf{masking}, \textbf{encoding}, \textbf{decoding}, and \textbf{reconstruction} steps. We would like to remind that the encoder and decoder of SMARTIES operate independently from the number of sensors seen during pretraining and also that we do not use sensor-specific encoders/decoders, since data from the different sensors are already projected in the common spectrum-aware space. This allows us to maintain a similar computational complexity to the vanilla MAE. 

\vspace{.1cm}\noindent\textbf{Masking}. Random masking is applied on both $\mathbf{T}_{a'}$ and $\mathbf{T}_{b'}$ with a ratio $R$. The remaining unmasked tokens are visible tokens: $\mathbf{T}^{vis}_{a'}$ and $\mathbf{T}^{vis}_{b'}$, both of size $(1-R)N_W N_H\times D$. They are fed into the encoder.

\vspace{.1cm}\noindent\textbf{Encoding}. We use a ViT architecture (e.g., ViT-B, ViT-L etc.), which processes tokens from $\mathbf{T}^{vis}_{a'}$ and $\mathbf{T}^{vis}_{b'}$ together with the special [CLS] token, and provides latent image representations for all of them. To encode the relative positioning of tokens, we use sinusoidal positional encodings (PE).

\vspace{.1cm}\noindent\textbf{Decoding}. The decoder receives a hybrid input consisting of the unmasked tokens and special [MASK] tokens leveraged by the positional encodings. The learnable [MASK] token, along with the positional encodings, is exploited to decode the masked patches at specific locations. 

\vspace{.1cm}\noindent\textbf{Reconstruction}. After the decoding phase, we reproject (i.e., reconstruct) the decoded image representations of the pair back to their original spectral channels, leading to reconstructed patches $\mathbf{\hat{P}}_{a} \in \mathbb{R}^{N_{W}\times N_{H}\times S^{2} C_{a}}$ and $\mathbf{\hat{P}}_{b} \in \mathbb{R}^{N_{W}\times N_{H}\times S^{2} C_{b}}$. Similar to the spectrum-aware projection, the spectrum-aware reconstruction uses different reprojection layers for different spectral ranges (i.e., bands). To this end, we first define a set of fully-connected reprojection layers $\mathcal{R}=\{r_1, r_2, ..., r_n\}$, where $r_i: \mathbb{R}^D \rightarrow \mathbb{R}^{S\times S}$ is the remapping function for the $i$th spectral range. There is one reprojection layer $r_i$ corresponding to each projection layer $f_i$. For the reconstruction of patch $p_a$ from the corresponding decoded token $t_{a}$, we use $t_{a}$ for each band of the patch, and apply the corresponding reprojection layers. The overall process is illustrated in the right panel of~\cref{fig:pipeline_spectrum}. Once this is applied for all the decoded tokens, the reconstructed masked patches $\mathbf{\hat{P}}^{mask}_{a'}$, $\mathbf{\hat{P}}^{mask}_{b'}$ are obtained.

To train our model in a self-supervised way, we use the Mean Squared Error (MSE) loss between the original masked patches $\mathbf{P}^{mask}_{a'}$ and the corresponding reconstructed ones $\mathbf{\hat{P}}^{mask}_{a'}$. For $\mathbf{I}_{a'}$, the reconstruction loss is:
\begin{equation}
    \mathcal{L}_{a'}=\frac{\sum(\mathbf{P}^{mask}_{a'}-\mathbf{\hat{P}}^{mask}_{a'})^2}{RN_W N_H}
\end{equation}
where the denominator denotes the number of masked tokens. The final reconstruction loss is computed over both mixed patches sets: $\mathcal{L}=\mathcal{L}_{a'} + \mathcal{L}_{b'}.$

\subsection{Downstream Transfer to Diverse Sensors}\label{sec:downstream} 
\begin{figure}[t]
  \centering
   \includegraphics[width=0.9\linewidth]{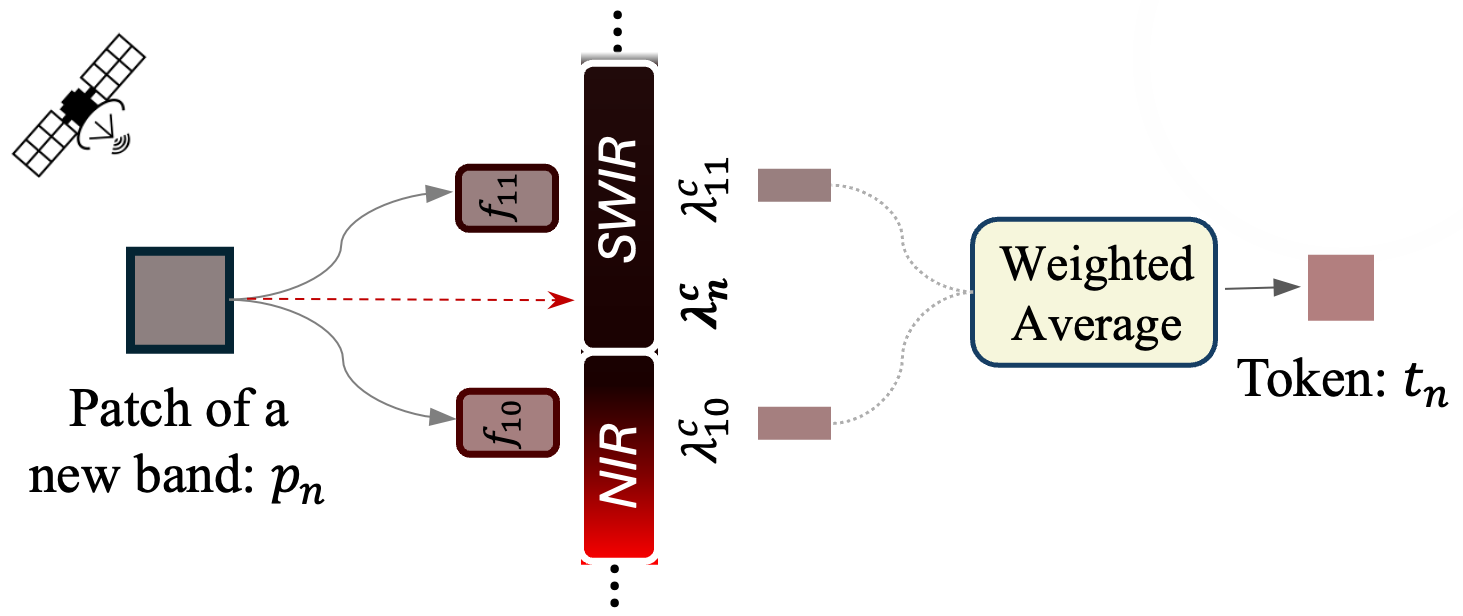}
   \caption{An example of downstream transfer to an unseen spectral band through interpolation. $\lambda^{c}_{10}$ and $\lambda^{c}_{11}$ denote the centre wavelength of the NIR and SWIR bands seen during pretraining; $\lambda^{c}_{n}$ denotes the centre wavelength of a new, unseen spectral band.}
   \label{fig:unseen_inference}
\end{figure}

After SMARTIES has been pretrained, the learned encoder and spectrum-aware projection layers can be used for various downstream tasks (\textit{e.g.}, classification, segmentation) with RS images from diverse sensors. For downstream transfer, there can be three main inference modes: (1) \textit{in-domain, seen sensor inference}, (2) \textit{open-domain, seen sensor inference}, and (3) \textit{open-domain, unseen sensor inference} (cf. \cref{fig:setups} in the supplementary material). For modes (1) and (2), since the sensors for the downstream task have been already seen during pretraining, one needs to select the relevant projection layers among the already learned ones for tokenization. On the contrary, for (3) the downstream transfer uses a sensor unseen during pretraining. This can be achieved in two ways. First, projection layers for the missing ranges can be easily learned during downstream transfer through full finetuning. However, this might not always be feasible, especially when full finetuning is costly to achieve. For such cases, as a second way, we apply \emph{interpolation} to unseen spectral ranges via a weighted average of the result of the closest projection layers. This is illustrated on a synthetic example in~\cref{fig:unseen_inference}, where an unseen sensor has a new band with a different spectral range than all sensors used during pretraining.
Since the central wavelength $\lambda^{c}_{n}$ of this new band falls between those of learnt layers $\lambda^{c}_{10}$ and $\lambda^{c}_{11}$, its token can be obtained by combining the corresponding projection layers $f_{10}$ and $f_{11}$, weighted by the normalized distances between the central wavelengths\footnote{Projection layer indices refer to \cref{fig:short-b}.}. However, we stress that this can: 1) only operate on the unseen ranges falling inside the minimum and maximum frequencies considered for pretraining; 2) not work for unseen regions out of the pretraining spectra (i.e., \emph{extrapolation}).

\section{Experiments}
\subsection{Pretraining Data}

To pretrain SMARTIES, we use paired images from: (1) the Functional Map of the World RGB dataset (fMoW-RGB)~\cite{fmow_2018_cvpr} and its Sentinel-2 counterpart fMoW-S2~\cite{satmae}; and (2) the BigEarthNet-MM~\cite{bigearthnetmm} dataset. To reduce the number of pretraining samples, we randomly selected 60K non-temporal fMoW pairs and 188K non-temporal BigEarthNet pairs, constituting a pretraining set of 496K images in total. This pretraining set is significantly smaller than those used in recent RS models (see \cref{supp:pt_efficiency_sec} in the supplementary material for a detailed comparison).

\vspace{.1cm}\noindent\textbf{fMoW} pairs represent various land use classes worldwide. fMoW-RGB~\cite{fmow_2018_cvpr} includes 470K submeter resolution RGB images of Maxar. fMoW-S2~\cite{satmae} includes over 882K Sentinel-2 images, containing 13 spectral bands with varying spatial resolutions for different bands (10m, 20m and 60m). The locations and temporal stamps of fMoW-S2 are mostly aligned with those of fMoW-RGB.

\vspace{.1cm}\noindent\textbf{BigEarthNet-MM} (BEN)~\cite{bigearthnetmm} includes over 590K pairs of multispectral and SAR images; each pair is acquired by Sentinel-2 and Sentinel-1 satellites on the same geographical area and associated with multi-labels. BEN-S2 images contain 12 spectral bands with 10m, 20m and 60m spatial resolutions, BEN-S1 images include dual-polarized information bands (VV and VH) with 10m spatial resolution. 

\vspace{.1cm}\noindent\textbf{Preprocessing for Data Harmonization} is achieved by first min-max image normalization with 1\% and 99\% percentile values, and then image standardization with mean and standard deviation values. This allows SMARTIES to be robust towards data distribution differences across multiple sensors (e.g., long-tailed distribution of 12 bit Sentinel-2 images \textit{vs.} short-tailed distribution of 8 bit RGB images).

\subsection{Experimental Setup}
To pretrain SMARTIES, we follow the same architectural choices and hyperparameters with the vanilla MAE~\cite{mae_2022_cvpr}, wherever possible. In detail, we pretrain two models with ViT-B and ViT-L~\cite{vit_2020_arxiv} backbones for 300 epochs, using the AdamW optimizer~\cite{adamw_2019_iclr} with the batch size of 2048 (distributed over 8 A100 GPUs) and the base learning rate of 1.5e-4. The masking ratio $R$ and the input size $W\times H$ are set to 75\% and 224$\times$224, respectively, as in the vanilla MAE. The mixup ratio is set to 50\%, while the size $D$ of spectrum-aware embeddings is set to 768 and 1024 for ViT-B and ViT-L, respectively (see \cref{supp:pt_sect} in the supplementary material for the details of pretraining). To assess the effectiveness of SMARTIES across different sensors, tasks, and scenarios, we consider the following experiments:
\begin{enumerate}
    \item[\textbullet] \textbf{Multispectral, Radar, and RGB Experiments.} We evaluate the performance of SMARTIES with various single-modal/multi-modal input (Multispectral, Radar, and RGB) to investigate its robustness across different sensors. More specifically, experiments under this group belong to the following two categories: \textit{in-domain, seen sensor} transfer, where performance is assessed on datasets seen during pretraining; \textit{open-domain, seen sensor} transfer generalization on datasets unseen during pretraining from known sensors.
    \item[\textbullet] \textbf{Unseen Sensor Transfer Experiments.} In this \textit{open-domain, unseen sensor} transfer, we evaluate the generalization capability of SMARTIES on new sensors not present in the pretraining data.
\end{enumerate}
We try to compare our models with all the existing models involving similar number of parameters and backbones on the modality they were designed for (e.g. Scale-MAE on RGB) to ensure fairness\footnote{For a broader comparison, see \cref{tab:efficiency_compare} in the supplementary material.}. Experimental comparison is conducted on the mostly used datasets and tasks of the previous studies, while following the same evaluation protocol: \textit{k}NN ($k=$ 20) classification, linear probing (LP), full finetuning (FT) and frozen backbone finetuning. We refer reader to the \cref{supp:eval_sect} in the supplementary material for the details of downstream transfer on each dataset and task.
\begin{table}
  \centering
  \small
  \renewcommand{\arraystretch}{0.87}
  \setlength\tabcolsep{3pt}
  \begin{tabular}{@{}lcccc@{}}
    \toprule
    \multirow{3}{*}{Method} & \multirow{3}{*}{Backbone} & \multicolumn{3}{c}{BEN 10\%} \\
    & & S1 (LP) & S2 (FT)& MM (LP) \\
    \midrule
    SeCo~\cite{seco} & RN-50 & \underline{69.9} & \textbf{82.6} & \underline{76.9}\\
    GASSL~\cite{gassl} & RN-50 & 66.1 & 80.2 & 73.2 \\
    CACo~\cite{caco} & RN-50 & \textbf{70.1} & \underline{81.3} & \textbf{78.5} \\\midrule
    SatMAE (S2)~\cite{satmae} & ViT-B & 68.4 & 85.9 & 77.8 \\
    GFM~\cite{gfm} & Swin-B & \underline{73.6} & 86.3 & \underline{82.0} \\
    SatLas (S2)~\cite{satlas} & Swin-B & 60.8 & 82.8 & 70.1 \\
    I-JEPA~\cite{ijpea} & ViT-B & \nap & 85.9 & \nap \\
    SpectralGPT~\cite{spectral_gpt} & ViT-B & 57.1 & 85.6 & 68.5 \\
    S2MAE~\cite{s2mae} & ViT-B & \nap & 85.6 & \nap \\
    msGFM~\cite{ms_gfm} & Swin-B & 67.5* & \underline{86.8} & \nap \\
    SMARTIES (Ours) & ViT-B & \textbf{78.9} & \textbf{86.9} & \textbf{85.4} \\ \midrule
    SatMAE (S2)~\cite{satmae} & ViT-L & 67.4 & 82.1 & 77.6 \\
    CROMA~\cite{croma} & ViT-B ($\times$2) & \underline{79.8} & \underline{87.6} & \underline{85.2} \\
    SpectralGPT~\cite{spectral_gpt} & ViT-L & \nap & 86.9 & \nap \\
    S2MAE~\cite{s2mae} & ViT-L & \nap & 86.5 & \nap \\
    SatMAE++ (S2)~\cite{satmae_pp} & ViT-L & 67.6 & 85.1 & 78.1 \\
    SMARTIES (Ours) & ViT-L & \textbf{80.5} & \textbf{87.7} & \textbf{86.7} \\ 
    \bottomrule
  \end{tabular}  
  \caption{BEN multi-label scene classification results (mAP) when linear probing (LP) or finetuning (FT) is applied with 10\% of the training set. {\nap} indicates \textit{not applicable} due to the lack of publicly available models. The highest results are written in bold, while the second best results are underlined. *We report FT result of msGFM as the LP result is not available in the original paper.}
  \label{tab:ben}
\end{table}

\subsection{Multispectral, Radar, and RGB Experiments}
\textbf{BigEarthNet.} In \cref{tab:ben}, we test the performance of SMARTIES on BigEarthNet-S1 (BEN-S1), BigEarthNet-S2 (BEN-S2), and multi-modal BigEarthNet (BEN-MM)~\cite{bigearthnetmm} by following the finetuning strategies mostly used in previous papers. By doing so, we assess the \textit{in-domain} representation ability of SMARTIES on datasets seen during pretraining, while focusing on multi-label RS scene classification task. The BEN setup, involving both S1 and S2, enables to study the ability of SMARTIES on handling known sensors variability. Results in Tab.~\ref{tab:ben} show that SMARTIES excels as a unified FM, demonstrating superior performance with diverse sensor inputs (single-modal/multi-modal) and distinguishing itself from previous methods that require sensor-specific pretraining. With various single-sensor input (columns S1 and S2 of Tab.~\ref{tab:ben}), SMARTIES consistently outperforms sensor-specific competitors on both BEN-S1 LP and BEN-S2 FT for ViT-B and ViT-L backbones. Since previous models are mainly customized for optical data (BEN-S2), most of them perform poorly under the BEN-S1 setting, composed of SAR data.
When considering multi-sensor fused inputs (column MM of Tab.~\ref{tab:ben}), SMARTIES outperforms previous state-of-the-art model CROMA (which was specifically pretrained with SAR-Optical image pairs), by significant 1.5\% mAP (ViT-L) through LP. In addition, the results on BEN-MM LP reflect the complementary benefits of leveraging multi-modal data using a single model: compared to LP with only SAR input (BEN-S1 LP), multi-modal input (BEN-MM LP) boosts the performance by 6.2\% mAP with our model.  
\begin{table}
    \small
  \centering
    \setlength\tabcolsep{18pt}
    \renewcommand{\arraystretch}{0.87}
  \begin{tabular}{@{}lcc@{}}
    \toprule
    Method & Backbone & LP / FT\\
    \midrule
    SeCo~\cite{seco} & RN-18 & \lna / \underline{93.1} \\
    GASSL~\cite{gassl} & RN-18 & \lna / 89.5\\
    SeCo~\cite{seco} & RN-50 & \underline{95.6} / \textbf{97.2} \\
    CACo~\cite{caco} & RN-50 & \textbf{95.9} / \rna \\\midrule
    SatMAE (S2)~\cite{satmae} & ViT-B & \underline{96.6} / 99.2\\
    I-JEPA~\cite{ijpea} & ViT-B & 95.6 / 99.2 \\
    SpectralGPT~\cite{spectral_gpt} & ViT-B & \lna / \underline{99.2} \\
    S2MAE~\cite{s2mae} & ViT-B & \lna / 99.2 \\
    SMARTIES (Ours) & ViT-B & \textbf{98.4} / \textbf{99.4} \\
    \midrule
    SatMAE (S2)~\cite{satmae} & ViT-L & \underline{97.7} / 99.0 \\
    SatMAE (RGB)~\cite{satmae} & ViT-L & 93.0 / 95.7 \\
    CROMA~\cite{croma} & ViT-B ($\times$2) & 97.6 / \underline{99.2} \\
    SatMAE++ (S2)~\cite{satmae_pp} & ViT-L & \lna / 99.0 \\
    SMARTIES (Ours) & ViT-L & \textbf{98.9} / \textbf{99.6} \\
    \bottomrule
  \end{tabular}
  \caption{Top-1 accuracy (\%) on EuroSAT. {\nap} indicates \textit{not available} results in the original papers.}
  \label{tab:eurosat}
\end{table}

\vspace{.1cm}\noindent\textbf{Downstream Transfer.} We test the \textit{open-domain, seen sensor} representation ability of SMARTIES on the downstream tasks of scene classification with datasets RESISC-45~\cite{resisc45_ieee_2017}, EuroSAT~\cite{eurosat_2019_jstars}, WHU-RS19~\cite{whurs19_2010_grsl} and UCMerced~\cite{ucmerced} which are not seen during pretraining. Remote Sensing Image Scene Classification (RESISC-45) is a very high-resolution RGB imagery dataset, which contains 31,500 images and 45 scene classes in total. EuroSAT is a multispectral dataset for scene classification, including 27K Sentinel-2 images with 10 classes. Results on the EuroSAT dataset are provided in~\cref{tab:eurosat}, where SMARTIES surpasses previous FMs both in LP (98.9\% \textit{vs.} 97.7\%) and FT (99.6\% \textit{vs.} 99.2\%). \cref{tab:resisc} shows the results on RESISC-45, where SMARTIES demonstrates highly competitive performance, even against models specifically trained on RGB data. Compared to previous methods, SMARTIES also showcases high data efficiency, achieving these results with only 496K images for pretraining, of which 60K RGB images represent only a small fraction (see \cref{supp:pt_efficiency_sec} in the supplementary material for a detailed analysis). These results in the \textit{open-domain, seen sensor} setup further demonstrate the remarkable generalization and scalability of SMARTIES.

\begin{table}
    \small
  \centering
    \setlength\tabcolsep{15pt}
    \renewcommand{\arraystretch}{0.87}
  \begin{tabular}{@{}lcc@{}}
    \toprule
    Method & Backbone & Top-1 Acc.\\
    \midrule
    MAE~\cite{mae_2022_cvpr} & ViT-L & 93.3\\
    SatMAE (RGB)~\cite{satmae} & ViT-L & 94.8 \\
    MCMAE~\cite{mcmae} & Conv ViT-L & 95.0 \\
    Scale-MAE~\cite{scale_mae} & ViT-L & 95.7\\
    SatMAE++ (RGB)~\cite{satmae_pp} & ViT-L & \textbf{97.5} \\
    SMARTIES (Ours) & ViT-L & \underline{95.8}\\
    \bottomrule
  \end{tabular}
  \caption{Top-1 accuracy (\%) of finetuning on RESISC-45.}
  \label{tab:resisc}
\end{table}
\begin{table}
  \centering
  \small
  \setlength\tabcolsep{6pt}
  \renewcommand{\arraystretch}{0.87}
  \begin{tabular}{@{}lccc@{}}
    \toprule
    Method & EuroSAT & WHU-RS19 & UCMerced \\
    \midrule
    SatMAE (RGB)~\cite{satmae} & 84.4 & 69.9 & 69.7 \\
    Scale-MAE~\cite{scale_mae} & 86.7 & 79.5 & \underline{75.0} \\
    Cross-Scale MAE~\cite{cross_scale_mae} & \underline{87.8} & \underline{79.8} & 74.5\\
    SMARTIES (Ours) & \textbf{93.7} & \textbf{80.4} & \textbf{77.0} \\ 
    \bottomrule
  \end{tabular}
  \caption{\textit{k}NN classification accuracies averaged over different scale ratios (12.5\%, 25\%, 50\%, 100\%).}
  \label{tab:multiscale}
\end{table}

\vspace{.1cm}\noindent\textbf{Multi-scale Transfer.} RS sensors exhibit pronounced differences in terms of spatial resolution. We mimic this variability in sensors characteristics by resizing images for different scale ratios (12.5\%, 25\%, 50\%, 100\%) for model evaluation. Experiments are performed on three datasets which are all unseen when pretraining: EuroSAT, WHU-RS19~\cite{whurs19_2010_grsl}, UCMerced~\cite{ucmerced} to assess model's ability on handling scale variances. WHU-RS19 and UCMerced are both very high-resolution RGB image datasets. More precisely, WHU-RS19 contains 19 typical RS scene classes and images with up to 0.5m spatial resolution, while UCMerced contains 21 land use classes of urban locations around the United States. \cref{tab:multiscale} presents the \textit{k}NN classification accuracies on this multi-scale setup for different FMs. Without scale-specific pretraining, SMARTIES significantly outperforms previous state-of-the-art FMs (SatMAE, Scale-MAE, and Cross-scale MAE), which are specifically designed to tackle multi-scale input, with an improvement 5.9\% on EuroSAT, 0.6\% on WHU-RS19, and 2.0\% on UCMerced. Results from this setup verifies the robustness of SMARTIES against scale variability. We argue that SMARTIES owes this feature to the multi-scale nature of the pretraining data.

\begin{table}
  \centering
  \small
  \setlength\tabcolsep{3.5pt}
  \renewcommand{\arraystretch}{0.87}
  \begin{tabular}{@{}lcccc@{}}
    \toprule
    Method & Backbone & BurnScars & DEN & SpaceNet7 \\ \midrule
    GFM~\cite{gfm} & Swin-B & 76.9 & 34.1 & 60.9 \\
    CROMA~\cite{croma} & ViT-B ($\times$2) & 81.8 & 38.3 & 59.9\\
    SenPa-MAE~\cite{senpa_mae} & ViT-B & 80.8 & 30.2 & 58.5\\
    DOFA~\cite{dofa_2024_arxiv} & ViT-B & 80.6 & \textbf{39.3} & \underline{61.8}\\
    TerraMindv1~\cite{terramind} & ViT-B & \underline{82.4} & 37.9 & 60.6 \\
    SMARTIES (Ours) & ViT-B & \textbf{82.8} & \underline{38.5} & \textbf{62.2} \\ 
    \bottomrule
  \end{tabular}
  \caption{Semantic segmentation results (mIoU) with frozen backbone UPerNet probing on BurnScars, DynamicEarthNet (DEN) and SpaceNet7, using the PANGAEA benchmark~\cite{pangaea}.}
  \label{tab:pangaea}
\end{table}
\begin{table}
  \centering
  \small
  \setlength\tabcolsep{9pt}
  \renewcommand{\arraystretch}{0.87}
  \begin{tabular}{@{}lcccc@{}}
    \toprule
     Method & Training & mIoU & Acc. & F1\\
    \midrule
    U-Net 2D~\cite{unet2d} & Scratch & 47.7 & 69.7 & 62.7\\
    DeepLapV3+~\cite{deeplabv3} & Scratch & \underline{48.5} & \underline{71.2} & \underline{63.2}\\
    SMARTIES (w/o PI) & Frozen & 35.4 & 55.8 & 50.6\\
    SMARTIES (Ours) & Frozen & \textbf{50.2} & \textbf{75.5} & \textbf{63.7}\\
    \bottomrule
  \end{tabular}
  \caption{Unseen sensor transfer of SMARTIES (ViT-B) for crop-type segmentation on SICKLE. PI: projection interpolation; Frozen: a segmentation head is finetuned with frozen backbone.}
  \label{tab:sickle}
\end{table}
\subsection{Unseen Sensor Transfer Experiments} 
The sensor-agnostic design of SMARTIES allows adapting to new sensors that were not present during pretraining (\cref{sec:downstream}). We verify this \textit{sensor transfer} capability in the \emph{open domain, unseen sensor} mode (see \cref{fig:setups} in the supplementary material) through frozen backbone semantic segmentation experiments on SICKLE~\cite{sickle_wacv_2024} for crop-type mapping, BurnScars~\cite{burnscars} for burn scars detection, DynamicEarthNet (DEN) for land-use and land-cover mapping~\cite{dynamicearthnet} and SpaceNet7~\cite{spacenet7} for building detection. When encoder and projection layers are kept frozen, the unseen sensor transfer of SMARTIES can be achieved via: 1) the classical naive approach of feeding the new sensor's bands to the closest projection layers; and 2) the proposed projection interpolation (cf. \cref{sec:downstream} and \cref{fig:unseen_inference}). We verify the former on the Planet images of DEN and SpaceNet7, and the Harmonized Landsat Sentinel-2 (HLS) images of BurnScars. Even though Planet and HLS sensors have been never seen during SMARTIES pretraining, wavelength ranges of their bands highly overlap with the pretraining spectra that allows to directly leverage the already learned projection layers. \cref{tab:pangaea} shows the results under a comparison with state-of-the-art multi-sensor FMs by using the evaluation protocol of the PANGAEA~\cite{pangaea} benchmark with frozen backbone UPerNet~\cite{upernet} probing. SMARTIES surpasses previous multi-sensor FMs on BurnScars and SpaceNet7 with a highly competitive performance on DEN. These results shows the success of SMARTIES for unseen sensor transfer in the case of overlap with pretraining spectra. SICKLE contains Landsat-8 satellite images, which are acquired by an optical sensor (OLI) and a thermal sensor (TIRS). TIRS sensor with its thermal infrared bands is characterized by wavelength ranges unseen during pretraining (i.e., non-overlapping with pretraining spectra). To test SMARTIES performance on the SICKLE’s Landsat-8 segmentation task by using both OLI and TIRS bands, we apply interpolation for the thermal infrared bands, and learn a single layer segmentation head to assess its unseen sensor transfer capability. Tab.~\ref{tab:sickle} shows a comparison of our model's performance with fully supervised models trained specifically on this dataset. SMARTIES with projection interpolation surpasses fully supervised models in all the metrics, and by a margin of 1.7\% in mIoU, even with frozen backbone parameters. By combining spectrum-aware projection layers (\cref{fig:pipeline_spectrum} and \cref{fig:unseen_inference}), SMARTIES achieves strong generalization and adaptability, showing its potential as a versatile FM for new sensor types without requiring additional sensor-specific finetuning.
\begin{figure}[t]
    \centering
        \begin{tikzpicture}[scale = 0.8]%
	\begin{axis}[
    legend columns=1,
    height=5cm,
    width=10.5cm,
    legend style={font=\scriptsize, at={(axis cs:250,93.4)},anchor=south west},
    grid=both,
    grid style={line width=.1pt, draw=gray!10},
    major grid style={line width=.2pt,draw=gray!50},
    minor x tick num=3,
    minor y tick num=4,
    xlabel= {\small Epoch},
    ylabel= {\small kNN Accuracy (\%)},
    xmin=40,xmax=310,
    ymin=93, ymax=95.8]
    \addplot+[name path=capacity,solid,color=Fuchsia,mark=pentagon,mark options={fill=white},line width=1pt] table [x=epoch, y=base, col sep=comma] {vit-epoch-comp.csv};\addlegendentry{ViT-B};
    \addplot+[name path=capacity,solid,color=Magenta,mark=*,mark options={fill=white},line width=1pt] table [x=epoch, y=large, col sep=comma] {vit-epoch-comp.csv};\addlegendentry{ViT-L};
	\end{axis}
    \end{tikzpicture}
    \caption{\textit{k}NN classification accuracy (\%) on EuroSAT versus pretraining epoch for ViT-B and ViT-L.}
\label{fig:sens}
\end{figure}
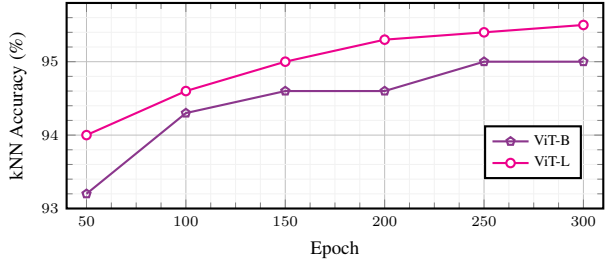
\subsection{Ablations}

\textbf{Model Size and Pretraining Epochs.} 
Our model is built based on the standard ViT backbones with the lightweight projectors, so that SMARTIES does not significantly increase parameter count compared to the standard MAE (+5.9M for ViT-L backbone).~\cref{fig:sens} presents the effects of ViT backbone and pretraining epochs for \textit{k}NN classification on EuroSAT. Overall, using ViT-L brings continuous improvements over ViT-B of about 0.5\% for the same epochs, showing the scalability of SMARTIES. 

\vspace{.1cm}\noindent\textbf{Cross-sensor Token Mixup.} 
We study the robustness of the learned representations by ablating the use of cross-sensor token mixup with models pretrained for 50 epochs (\cref{tab:ablation_mixup}). With mixup, the model's performance in processing multi-modal data is significantly enhanced: the LP performance increases by 2.2\% on BEN-MM. We argue that mixup adds variability in the input data that helps to prevent overfitting, even with BEN only pretraining.

\vspace{.1cm}\noindent\textbf{Fusion Strategies for Multi-modal Input.} We also examine the effects of feature fusion strategies for the multi-modal downstream transfer. As shown in Tab.~\ref{tab:ablation_fusion}, we design three fusion strategies to adapt to multi-modal inputs: \textit{Image Stacking} denotes a strategy directly stacking images from different modalities as input to the model; \textit{Feature Concatenation} means concatenating features obtained by the encoder for different modalities; finally, \textit{Mixup Concatenation} is our proposed strategy, which concatenates features extracted from the encoder from different modalities after applying mixup with spectrum-aware projections. An illustration of these strategies is given in \cref{fig:pipeline_inference} in the supplementary material. Results in~\cref{tab:ablation_fusion} show that Mixup Concatenation achieves the highest mAP in both 1\% and 10\% training data of BEN-MM for LP.

\begin{table}
  \centering
  \small
  \setlength\tabcolsep{14pt}
  \renewcommand{\arraystretch}{0.87}
  \begin{tabular}{@{}lcc@{}}
    \toprule
     Method & Backbone & Acc.\\
    \midrule
    SMARTIES (w/o mixup) & ViT-B & 91.0\\
    SMARTIES (w mixup, BEN only) & ViT-B & \underline{91.1}\\
    SMARTIES (w mixup) & ViT-B & \textbf{93.2}\\
    \bottomrule
  \end{tabular}
  \caption{\textit{k}NN classification accuracy (\%) on EuroSAT under different pretraining settings (pretraining epochs are set to 50).}
  \label{tab:ablation_mixup}
\end{table}
\begin{table}
  \centering
  \small
  \setlength\tabcolsep{14pt}
  \renewcommand{\arraystretch}{0.87}
  \begin{tabular}{@{}lccc@{}}
    \toprule
     Strategy & Backbone & 1\% & 10\%\\
    \midrule
    Image Stacking & ViT-L & 75.9 & 83.1\\
    Feature Concatenation & ViT-L & \underline{77.0} & \underline{84.7}\\
    Mixup Concatenation & ViT-L & \textbf{79.2} & \textbf{86.7}\\
    \bottomrule
  \end{tabular}
  \caption{mAP on BEN-MM for linear probing under different feature fusion strategies with 1\% and 10\% of the training sets.}
  \label{tab:ablation_fusion}
\end{table}

\section{Conclusion}
The variety of sensors that acquire a continuous stream of information characterizing the Earth's surface makes RS data multi-modal by nature. In this paper, we introduce SMARTIES, a unified and versatile foundation model that achieves sensor-agnostic representations by projecting diverse sensory data into shared spectrum-aware space and training with the masked reconstruction objective with cross-sensor token mixup. 
This strategy has the advantage of 
seamlessly handling diverse sensory inputs at pretraining, exhibiting scalability to RS sensors characterized by different spectral properties. SMARTIES can learn transferable representations not only for pretraining sensors but also for unseen ones, demonstrating unprecedented generalization capabilities. SMARTIES outperforms existing foundation models for RS in ten datasets on both single-modal and multi-modal tasks, including experiments testing the downstream transfer to the sensor never seen during pretraining. 
Unifying multi-sensor RS image interpretation with a single foundation model has the potential to leverage the synergistic advantages of different sensors for Earth observation, while eliminating the need for isolated efforts in training sensor-specific models. SMARTIES is one of the first steps in that direction, and extensions to the temporal domain and to more diversified downstream tasks will be our future efforts toward unified, physics-inspired foundation models for RS.

\section*{Acknowledgment}
This work is supported by the European Space Agency (ESA) through the Discovery and Preparation Program, and is part of the project Toward a Foundation Model for Multi-Sensor Earth Observation Data with Language Semantics.

{
\small
    \bibliographystyle{ieeenat_fullname}
    \bibliography{main}

\begin{thebibliography}{48}
\providecommand{\natexlab}[1]{#1}
\providecommand{\url}[1]{\texttt{#1}}
\expandafter\ifx\csname urlstyle\endcsname\relax
  \providecommand{\doi}[1]{doi: #1}\else
  \providecommand{\doi}{doi: \begingroup \urlstyle{rm}\Url}\fi

\bibitem[Assran et~al.(2023)Assran, Duval, Misra, Bojanowski, Vincent, Rabbat, LeCun, and Ballas]{ijpea}
Mahmoud Assran, Quentin Duval, Ishan Misra, Piotr Bojanowski, Pascal Vincent, Michael Rabbat, Yann LeCun, and Nicolas Ballas.
\newblock Self-supervised learning from images with a joint-embedding predictive architecture.
\newblock In \emph{IEEE/CVF Conf. Comput. Vis. Pattern Recog. (CVPR)}, pages 15619--15629, 2023.

\bibitem[Ayush et~al.(2021)Ayush, Uzkent, Meng, Tanmay, Burke, Lobell, and Ermon]{gassl}
Kumar Ayush, Burak Uzkent, Chenlin Meng, Kumar Tanmay, Marshall Burke, David Lobell, and Stefano Ermon.
\newblock Geography-aware self-supervised learning.
\newblock In \emph{Int. Conf. Comput. Vis. (ICCV)}, pages 10181--10190, 2021.

\bibitem[Bachmann et~al.(2022)Bachmann, Mizrahi, Atanov, and Zamir]{multimae_2022_eccv}
Roman Bachmann, David Mizrahi, Andrei Atanov, and Amir Zamir.
\newblock Multimae: Multi-modal multi-task masked autoencoders.
\newblock In \emph{Eur. Conf. Comput. Vis. (ECCV)}, pages 348--367. Springer, 2022.

\bibitem[Bastani et~al.(2023)Bastani, Wolters, Gupta, Ferdinando, and Kembhavi]{satlas}
Favyen Bastani, Piper Wolters, Ritwik Gupta, Joe Ferdinando, and Aniruddha Kembhavi.
\newblock Satlaspretrain: A large-scale dataset for remote sensing image understanding.
\newblock In \emph{Int. Conf. Comput. Vis. (ICCV)}, pages 16772--16782, 2023.

\bibitem[Chen et~al.(2018)Chen, Zhu, Papandreou, Schroff, and Adam]{deeplabv3}
Liang-Chieh Chen, Yukun Zhu, George Papandreou, Florian Schroff, and Hartwig Adam.
\newblock Encoder-decoder with atrous separable convolution for semantic image segmentation.
\newblock In \emph{Eur. Conf. Comput. Vis. (ECCV)}, pages 801--818, 2018.

\bibitem[Cheng et~al.(2017)Cheng, Han, and Lu]{resisc45_ieee_2017}
Gong Cheng, Junwei Han, and Xiaoqiang Lu.
\newblock Remote sensing image scene classification: Benchmark and state of the art.
\newblock \emph{Proceedings of the IEEE}, 105\penalty0 (10):\penalty0 1865--1883, 2017.

\bibitem[Chi et~al.(2016)Chi, Plaza, Benediktsson, Sun, Shen, and Zhu]{rsbigdata_ieee}
Mingmin Chi, Antonio Plaza, Jón~Atli Benediktsson, Zhongyi Sun, Jinsheng Shen, and Yangyong Zhu.
\newblock Big data for remote sensing: Challenges and opportunities.
\newblock \emph{Proceedings of the IEEE}, 104\penalty0 (11):\penalty0 2207--2219, 2016.

\bibitem[Christie et~al.(2018)Christie, Fendley, Wilson, and Mukherjee]{fmow_2018_cvpr}
Gordon Christie, Neil Fendley, James Wilson, and Ryan Mukherjee.
\newblock Functional map of the world.
\newblock In \emph{IEEE/CVF Conf. Comput. Vis. Pattern Recog. (CVPR)}, pages 6172--6180, 2018.

\bibitem[Cong et~al.(2022)Cong, Khanna, Meng, Liu, Rozi, He, Burke, Lobell, and Ermon]{satmae}
Yezhen Cong, Samar Khanna, Chenlin Meng, Patrick Liu, Erik Rozi, Yutong He, Marshall Burke, David~B. Lobell, and Stefano Ermon.
\newblock Satmae: Pre-training transformers for temporal and multi-spectral satellite imagery.
\newblock In \emph{Adv. Neural Inform. Process. Syst. (NeurIPS)}, pages 197--211, 2022.

\bibitem[Dai and Yang(2010)]{whurs19_2010_grsl}
Dengxin Dai and Wen Yang.
\newblock Satellite image classification via two-layer sparse coding with biased image representation.
\newblock \emph{IEEE Geosci. Remote Sens. Lett. (GRSL)}, 8\penalty0 (1):\penalty0 173--176, 2010.

\bibitem[Dosovitskiy et~al.(2021)Dosovitskiy, Beyer, Kolesnikov, Weissenborn, Zhai, Unterthiner, Dehghani, Minderer, Heigold, Gelly, Uszkoreit, and Houlsby]{vit_2020_arxiv}
Alexey Dosovitskiy, Lucas Beyer, Alexander Kolesnikov, Dirk Weissenborn, Xiaohua Zhai, Thomas Unterthiner, Mostafa Dehghani, Matthias Minderer, Georg Heigold, Sylvain Gelly, Jakob Uszkoreit, and Neil Houlsby.
\newblock An image is worth 16x16 words: Transformers for image recognition at scale.
\newblock In \emph{Int. Conf. Learn. Represent. (ICLR)}, 2021.

\bibitem[Fuller et~al.(2023)Fuller, Millard, and Green]{croma}
Anthony Fuller, Koreen Millard, and James~R. Green.
\newblock {CROMA}: Remote sensing representations with contrastive radar-optical masked autoencoders.
\newblock In \emph{Adv. Neural Inform. Process. Syst. (NeurIPS)}, pages 5506--5538, 2023.

\bibitem[Gao et~al.(2022)Gao, Ma, Li, Lin, Dai, and Qiao]{mcmae}
Peng Gao, Teli Ma, Hongsheng Li, Ziyi Lin, Jifeng Dai, and Yu Qiao.
\newblock {MCMAE}: Masked convolution meets masked autoencoders.
\newblock In \emph{Adv. Neural Inform. Process. Syst. (NeurIPS)}, pages 35632--35644, 2022.

\bibitem[G{\'o}mez-Chova et~al.(2015)G{\'o}mez-Chova, Tuia, Moser, and Camps-Valls]{multimodalrs_survey_ieee}
Luis G{\'o}mez-Chova, Devis Tuia, Gabriele Moser, and Gustau Camps-Valls.
\newblock Multimodal classification of remote sensing images: A review and future directions.
\newblock \emph{Proceedings of the IEEE}, 103\penalty0 (9):\penalty0 1560--1584, 2015.

\bibitem[Guo et~al.(2024)Guo, Lao, Dang, Zhang, Yu, Ru, Zhong, Huang, et~al.]{skysense}
Xin Guo, Jiangwei Lao, Bo Dang, Yingying Zhang, Lei Yu, Lixiang Ru, Liheng Zhong, Ziyuan Huang, et~al.
\newblock {SkySense}: A multi-modal remote sensing foundation model towards universal interpretation for earth observation imagery.
\newblock In \emph{IEEE/CVF Conf. Comput. Vis. Pattern Recog. (CVPR)}, pages 27662--27673, 2024.

\bibitem[Han et~al.(2024)Han, Zhang, Shi, and Reichstein]{ms_gfm}
Boran Han, Shuai Zhang, Xingjian Shi, and Markus Reichstein.
\newblock Bridging remote sensors with multisensor geospatial foundation models.
\newblock In \emph{IEEE/CVF Conf. Comput. Vis. Pattern Recog. (CVPR)}, pages 27852--27862, 2024.

\bibitem[He et~al.(2022)He, Chen, Xie, Li, Doll{\'a}r, and Girshick]{mae_2022_cvpr}
Kaiming He, Xinlei Chen, Saining Xie, Yanghao Li, Piotr Doll{\'a}r, and Ross Girshick.
\newblock Masked autoencoders are scalable vision learners.
\newblock In \emph{IEEE/CVF Conf. Comput. Vis. Pattern Recog. (CVPR)}, pages 16000--16009, 2022.

\bibitem[Helber et~al.(2019)Helber, Bischke, Dengel, and Borth]{eurosat_2019_jstars}
Patrick Helber, Benjamin Bischke, Andreas Dengel, and Damian Borth.
\newblock Eurosat: A novel dataset and deep learning benchmark for land use and land cover classification.
\newblock \emph{IEEE J. Sel. Top. Appl. Earth Obs. Remote Sens. (JSTARS)}, 12\penalty0 (7):\penalty0 2217--2226, 2019.

\bibitem[Hong et~al.(2024)Hong, Zhang, Li, Li, Li, Yao, Yokoya, Li, et~al.]{spectral_gpt}
Danfeng Hong, Bing Zhang, Xuyang Li, Yuxuan Li, Chenyu Li, Jing Yao, Naoto Yokoya, Hao Li, et~al.
\newblock {SpectralGPT}: Spectral remote sensing foundation model.
\newblock \emph{IEEE Trans. Pattern Anal. Mach. Intell. (TPAMI)}, 46\penalty0 (8):\penalty0 5227--5244, 2024.

\bibitem[Jakubik et~al.(2023)Jakubik, Roy, Phillips, Fraccaro, Godwin, Zadrozny, Szwarcman, Gomes, et~al.]{burnscars}
Johannes Jakubik, Sujit Roy, C.~E. Phillips, Paolo Fraccaro, Denys Godwin, Bianca Zadrozny, Daniela Szwarcman, Carlos Gomes, et~al.
\newblock Foundation models for generalist geospatial artificial intelligence.
\newblock \emph{arXiv preprint arXiv:2310.18660}, 2023.

\bibitem[Jakubik et~al.(2025)Jakubik, Yang, Blumenstiel, Scheurer, Sedona, Maurogiovanni, Bosmans, Dionelis, et~al.]{terramind}
Johannes Jakubik, Felix Yang, Benedikt Blumenstiel, Erik Scheurer, Rocco Sedona, Stefano Maurogiovanni, Jente Bosmans, Nikolaos Dionelis, et~al.
\newblock {TerraMind}: Large-scale generative multimodality for earth observation.
\newblock \emph{arXiv preprint arXiv:2504.11171}, 2025.

\bibitem[Leenstra et~al.(2021)Leenstra, Marcos, Bovolo, and {Tuia}]{Len20}
Marrit Leenstra, Diego Marcos, Francesca Bovolo, and {Devis} {Tuia}.
\newblock Self-supervised pre-training enhances change detection in {S}entinel-2 images.
\newblock In \emph{Int. Conf. Pattern Recog. (ICPR), Workshop Pattern Recog. in Remote Sens.}, pages 578--590, 2021.

\bibitem[Li et~al.(2024)Li, Hong, and Chanussot]{s2mae}
Xuyang Li, Danfeng Hong, and Jocelyn Chanussot.
\newblock S2mae: A spatial-spectral pretraining foundation model for spectral remote sensing data.
\newblock In \emph{IEEE/CVF Conf. Comput. Vis. Pattern Recog. (CVPR)}, pages 27696--27705, 2024.

\bibitem[Li et~al.(2022)Li, Mao, Girshick, and He]{Li:2022}
Yanghao Li, Hanzi Mao, Ross Girshick, and Kaiming He.
\newblock Exploring plain vision transformer backbones for object detection.
\newblock \emph{arXiv preprint arXiv:2203.16527}, 2022.

\bibitem[Loshchilov and Hutter(2019)]{adamw_2019_iclr}
Ilya Loshchilov and Frank Hutter.
\newblock Decoupled weight decay regularization.
\newblock In \emph{Int. Conf. Learn. Represent. (ICLR)}, 2019.

\bibitem[Ma\~nas et~al.(2021)Ma\~nas, Lacoste, Gir\'o-i Nieto, Vazquez, and Rodr{\'\i}guez]{seco}
Oscar Ma\~nas, Alexandre Lacoste, Xavier Gir\'o-i Nieto, David Vazquez, and Pau Rodr{\'\i}guez.
\newblock Seasonal contrast: Unsupervised pre-training from uncurated remote sensing data.
\newblock In \emph{Int. Conf. Comput. Vis. (ICCV)}, pages 9414--9423, 2021.

\bibitem[Mall et~al.(2023)Mall, Hariharan, and Bala]{caco}
Utkarsh Mall, Bharath Hariharan, and Kavita Bala.
\newblock Change-aware sampling and contrastive learning for satellite images.
\newblock In \emph{IEEE/CVF Conf. Comput. Vis. Pattern Recog. (CVPR)}, pages 5261--5270, 2023.

\bibitem[Marsocci et~al.(2024)Marsocci, Jia, Bellier, Kerekes, Zeng, Hafner, Gerard, Brune, et~al.]{pangaea}
Valerio Marsocci, Yuru Jia, Georges~Le Bellier, David Kerekes, Liang Zeng, Sebastian Hafner, Sebastian Gerard, Eric Brune, et~al.
\newblock {PANGAEA}: A global and inclusive benchmark for geospatial foundation models.
\newblock \emph{arXiv preprint 2412.04204}, 2024.

\bibitem[Mendieta et~al.(2023)Mendieta, Han, Shi, Zhu, and Chen]{gfm}
Mat{\'\i}as Mendieta, Boran Han, Xingjian Shi, Yi Zhu, and Chen Chen.
\newblock Towards geospatial foundation models via continual pretraining.
\newblock In \emph{Int. Conf. Comput. Vis. (ICCV)}, pages 16806--16816, 2023.

\bibitem[Mizrahi et~al.(2023)Mizrahi, Bachmann, Kar, Yeo, Gao, Dehghan, and Zamir]{4m_nips_2023}
David Mizrahi, Roman Bachmann, Oguzhan Kar, Teresa Yeo, Mingfei Gao, Afshin Dehghan, and Amir Zamir.
\newblock {4M}: Massively multimodal masked modeling.
\newblock \emph{Adv. Neural Inform. Process. Syst. (NeurIPS)}, pages 58363--58408, 2023.

\bibitem[Nedungadi et~al.(2024)Nedungadi, Kariryaa, Oehmcke, Belongie, Igel, and Lang]{mmearth_eccv_2024}
Vishal Nedungadi, Ankit Kariryaa, Stefan Oehmcke, Serge Belongie, Christian Igel, and Nico Lang.
\newblock Mmearth: Exploring multi-modal pretext tasks for geospatial representation learning.
\newblock In \emph{Eur. Conf. Comput. Vis. (ECCV)}, pages 164--182, 2024.

\bibitem[Noman et~al.(2024)Noman, Naseer, Cholakkal, Anwar, Khan, and Khan]{satmae_pp}
Mubashir Noman, Muzammal Naseer, Hisham Cholakkal, Rao~Muhammad Anwar, Salman Khan, and Fahad~Shahbaz Khan.
\newblock Rethinking transformers pre-training for multi-spectral satellite imagery.
\newblock In \emph{IEEE/CVF Conf. Comput. Vis. Pattern Recog. (CVPR)}, pages 27811--27819, 2024.

\bibitem[Prexl and Schmitt(2024)]{senpa_mae}
Jonathan Prexl and Michael Schmitt.
\newblock {SenPa-MAE}: Sensor parameter aware masked autoencoder for multi-satellite self-supervised pretraining.
\newblock In \emph{German Conf. Pattern Recog. (GCPR)}, 2024.

\bibitem[Reed et~al.(2023)Reed, Gupta, Li, Brockman, Funk, Clipp, Keutzer, Candido, Uyttendaele, and Darrell]{scale_mae}
Colorado~J Reed, Ritwik Gupta, Shufan Li, Sarah Brockman, Christopher Funk, Brian Clipp, Kurt Keutzer, Salvatore Candido, Matt Uyttendaele, and Trevor Darrell.
\newblock Scale-mae: A scale-aware masked autoencoder for multiscale geospatial representation learning.
\newblock In \emph{Int. Conf. Comput. Vis. (ICCV)}, pages 4088--4099, 2023.

\bibitem[Ronneberger et~al.(2015)Ronneberger, Fischer, and Brox]{unet2d}
Olaf Ronneberger, Philipp Fischer, and Thomas Brox.
\newblock {U-Net}: Convolutional networks for biomedical image segmentation.
\newblock In \emph{Int. Conf. Medical Image Comput. Computer-assisted Intervention (MICCAI)}, pages 234--241. Springer, 2015.

\bibitem[Sainte Fare~Garnot and Landrieu(2020)]{ltae}
Vivien Sainte Fare~Garnot and Loic Landrieu.
\newblock Lightweight temporal self-attention for classifying satellite images time series.
\newblock \emph{arXiv preprint arXiv:2007.00586}, 2020.

\bibitem[Sani et~al.(2024)Sani, Mahato, Saini, Agarwal, Devshali, Anand, Arora, and Jayaraman]{sickle_wacv_2024}
Depanshu Sani, Sandeep Mahato, Sourabh Saini, Harsh~Kumar Agarwal, Charu~Chandra Devshali, Saket Anand, Gaurav Arora, and Thiagarajan Jayaraman.
\newblock {SICKLE}: A multi-sensor satellite imagery dataset annotated with multiple key cropping parameters.
\newblock In \emph{IEEE/CVF Winter Conf. on App. of Comput. Vis. (WACV)}, pages 5995--6004, 2024.

\bibitem[Scheibenreif et~al.(2022)Scheibenreif, Mommert, and Borth]{scheibenreif2022contrastive}
Linus Scheibenreif, Michael Mommert, and Damian Borth.
\newblock Contrastive self-supervised data fusion for satellite imagery.
\newblock \emph{ISPRS Ann. Photogramm. Remote Sens. Spatial Inf. Sci.}, 3:\penalty0 705--711, 2022.

\bibitem[Sumbul et~al.(2021)Sumbul, de~Wall, Kreuziger, Marcelino, Costa, Benevides, Caetano, Demir, and Markl]{bigearthnetmm}
Gencer Sumbul, Arne de Wall, Tristan Kreuziger, Filipe Marcelino, Hugo Costa, Pedro Benevides, Mário Caetano, Begüm Demir, and Volker Markl.
\newblock {BigEarthNet-MM}: A large scale multi-modal multi-label benchmark archive for remote sensing image classification and retrieval.
\newblock \emph{IEEE Geosci. Remote Sens. Magazine (GRSM)}, 9\penalty0 (3):\penalty0 174--180, 2021.

\bibitem[Sumbul et~al.(2022)Sumbul, Müller, and Demir]{Sumbul_2022}
Gencer Sumbul, Markus Müller, and Begüm Demir.
\newblock A novel self-supervised cross-modal image retrieval method in remote sensing.
\newblock In \emph{IEEE Int. Conf. Image Process. (ICIP)}, pages 2426--2430, 2022.

\bibitem[Tang et~al.(2023)Tang, Cozma, Georgiou, and Qi]{cross_scale_mae}
Maofeng Tang, Andrei~Liviu Cozma, Konstantinos Georgiou, and Hairong Qi.
\newblock Cross-scale {MAE}: A tale of multiscale exploitation in remote sensing.
\newblock In \emph{Adv. Neural Inform. Process. Syst. (NeurIPS)}, pages 20054--20066, 2023.

\bibitem[Toker et~al.(2022)Toker, Kondmann, Weber, Eisenberger, Andres, Hu, Hoderlein, Senaras, et~al.]{dynamicearthnet}
Aysim Toker, Lukas Kondmann, Mark Weber, Marvin Eisenberger, Camero Andres, Jingliang Hu, Ariadna Hoderlein, Caglar Senaras, et~al.
\newblock {DynamicEarthNet}: Daily multi-spectral satellite dataset for semantic change segmentation.
\newblock In \emph{IEEE/CVF Conf. Comput. Vis. Pattern Recog. (CVPR)}, 2022.

\bibitem[{Tuia} et~al.(2021){Tuia}, Roscher, Wegner, Jacobs, Zhu, and Camps-Valls]{Tui20grsm}
{Devis} {Tuia}, Ribana Roscher, Jan~Dirk Wegner, Nathan Jacobs, Xiao~Xiang Zhu, and Gustua Camps-Valls.
\newblock Towards a collective agenda on {AI} for earth science data analysis.
\newblock \emph{IEEE Geosci. Remote Sens. Magazine (GRSM)}, 9\penalty0 (2):\penalty0 88--104, 2021.

\bibitem[Van~Etten et~al.(2021)Van~Etten, Hogan, Manso, Shermeyer, Weir, and Lewis]{spacenet7}
Adam Van~Etten, Daniel Hogan, Jesus~Martinez Manso, Jacob Shermeyer, Nicholas Weir, and Ryan Lewis.
\newblock The multi-temporal urban development spacenet dataset.
\newblock In \emph{IEEE/CVF Conf. Comput. Vis. Pattern Recog. (CVPR)}, pages 6394--6403, 2021.

\bibitem[Wang et~al.(2022)Wang, Albrecht, Braham, Mou, and Zhu]{wang2022self}
Yi Wang, Conrad~M Albrecht, Nassim Ait~Ali Braham, Lichao Mou, and Xiao~Xiang Zhu.
\newblock Self-supervised learning in remote sensing: A review.
\newblock \emph{IEEE Geosci. Remote Sens. Magazine (GRSM)}, 10\penalty0 (4):\penalty0 213--247, 2022.

\bibitem[Xiao et~al.(2018)Xiao, Liu, Zhou, Jiang, and Sun]{upernet}
Tete Xiao, Yingcheng Liu, Bolei Zhou, Yuning Jiang, and Jian Sun.
\newblock Unified perceptual parsing for scene understanding.
\newblock In \emph{Eur. Conf. Comput. Vis. (ECCV)}, 2018.

\bibitem[Xiong et~al.(2024)Xiong, Wang, Zhang, Stewart, Hanna, Borth, Papoutsis, Saux, Camps-Valls, and Zhu]{dofa_2024_arxiv}
Zhitong Xiong, Yi Wang, Fahong Zhang, Adam~J Stewart, Jo{\"e}lle Hanna, Damian Borth, Ioannis Papoutsis, Bertrand~Le Saux, Gustau Camps-Valls, and Xiao~Xiang Zhu.
\newblock Neural plasticity-inspired multimodal foundation model for earth observation.
\newblock \emph{arXiv preprint arXiv:2403.15356}, 2024.

\bibitem[Yang and Newsam(2010)]{ucmerced}
Yi Yang and Shawn Newsam.
\newblock Bag-of-visual-words and spatial extensions for land-use classification.
\newblock In \emph{SIGSPATIAL Int. Conf. Advances in Geographic Inf. Systems}, page 270–279, 2010.

\end{thebibliography}
}

\clearpage 
\setcounter{section}{0}
\setcounter{figure}{0}
\setcounter{table}{0}
\renewcommand\thesection{S\arabic{section}}
\renewcommand\thefigure{S\arabic{figure}}
\renewcommand\thetable{S\arabic{table}}    
\setcounter{page}{1}

\maketitlesupplementary

In the supplementary material, we provide detailed information for pretraining and evaluation across different datasets. Besides, we provide additional analyses, including the pretraining efficiency of SMARTIES, more ablation results on the use of pretraining data and projection extrapolation to unseen spectral ranges. Our code and pretrained models are available at \url{https://gsumbul.github.io/smarties}.

\section{Implementation}
\label{supp:imp_sect}
\subsection{Spectrum-Aware Projection Layers}
\label{supp:projection_sect}
We provide the details for our projection layers $f_i$ in~\cref{tab:projection}. The spectral range of each layer is defined according to the bands of different sensors. Specifically, $f_1$ to $f_{12}$ follows the bands in Sentinel-2, $f_{13}$ to $f_{15}$ are based on RGB images from Maxar, and $f_{16}$, $f_{17}$ corresponds to Sentinel-1. Each reprojection layer $r_i$ takes charge of the same wavelength range as its corresponding projection layer $f_i$.
\begin{table}[b]
  \centering
    \setlength\tabcolsep{4pt}
  \begin{tabular}{@{}lcccc@{}}
    \toprule
     Sensor & Dataset & Layer & Band & Wavelength (nm)\\
    \midrule
     \multirow{12}{*}{S2} & \multirow{12}{*}{\shortstack[l]{BEN-S2\\fMoW-S2}} & $f_1$ & B01 & $422$ - $463$ \\
      & & $f_2$ & B02 & $427$ - $558$ \\
      & & $f_3$ & B03 & $524$ - $595$ \\
      & &$f_4$ & B04 & $634$ - $696$ \\
      & & $f_5$ & B05 & $689$ - $719$ \\
      & & $f_6$ & B06 & $726$ - $755$ \\
      & & $f_7$ & B07 & $761$ - $802$ \\
      & & $f_8$ & B08 & $728$ - $938$ \\
      & & $f_9$ & B8A & $843$ - $886$  \\
      & & $f_{10}$ &  B09 & $923$ - $964$ \\
      & & $f_{11}$ &  B11 & $1516$ - $1704$\\
      & & $f_{12}$ &  B12 & $2002$ - $2376$ \\
     \midrule
     \multirow{3}{*}{Maxar} & \multirow{3}{*}{fMoW-RGB} & $f_{13}$ & Blue & $430$ - $545$\\
      & & $f_{14}$ & Green & $466$ - $620$\\
      & & $f_{15}$  & Red & $590$ - $710$ \\
     \midrule
      \multirow{2}{*}{S1} & \multirow{2}{*}{BEN-S1} & $f_{16}$ & VV & $5.5\!\times\!\! 10^7$ - $5.6\!\times\!\! 10^7$\\
      & & $f_{17}$ & VH & $5.5\!\times\!\! 10^7$ - $5.6\!\times\!\! 10^7$\\
    \bottomrule
  \end{tabular}
  \caption{Spectral ranges for the projection layers used in SMARTIES pretraining. S2 denotes Sentinel-2, S1 denotes Sentinel-1, BEN is the abbreviation for BigEarthNet.}
  \label{tab:projection}
\end{table}

\begin{figure}[t]
  \centering
   \includegraphics[width=0.99\linewidth]{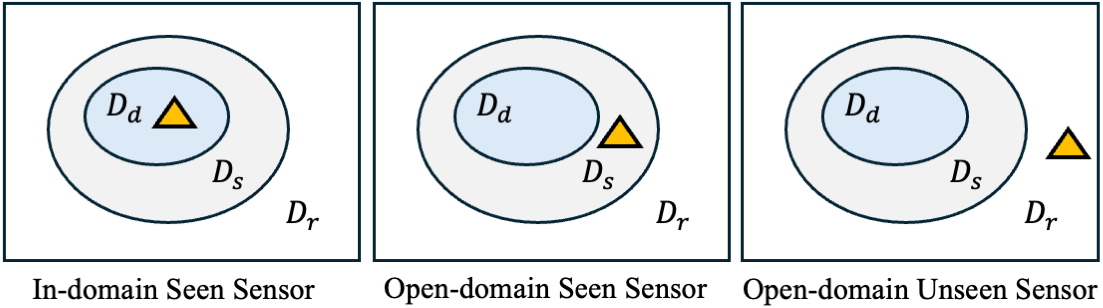}
   \caption{Different inference modes for downstream transfer to diverse sensors: (1) \emph{in-domain, seen sensor}: transfer in the domain $D_d$ of the same datasets seen during pretraining, (2) \emph{open domain, seen sensor}: transfer in the domain $D_s$ of new task, observed by the same sensors used during pretraining and (3) \emph{open domain, unseen sensor}: transfer in the domain $D_r$ of new tasks observed by any sensor. The yellow triangle denotes the position of each inference mode. From $D_d$ to $D_r$, an increasing degree of generalization is required.}
   \label{fig:setups}
\end{figure}
\subsection{Pretraining}
\label{supp:pt_sect}
We pretrain two versions of SMARTIES by using ViT-B and ViT-L~\cite{vit_2020_arxiv} backbones: SMARTIES (ViT-B) and SMARTIES (ViT-L), while we use the same decoders with the vanilla MAE~\cite{mae_2022_cvpr}. For both versions, we pretrain for 300 epochs, using AdamW optimizer~\cite{adamw_2019_iclr} ($\beta_1=0.9$, $\beta_1=0.95$ and weight decay of 0.05) and mixed precision (FP16) with the batch size of 2048 (distributed over 8 A100 GPUs), the base learning rate of 1.5e-4, warmup of 20 epochs and cooldown by half-cosine decay schedule. For data augmentation, we randomly apply vertical flipping, horizontal flipping and rotation in order. After this, by randomly sampling scale parameter between 0.25 and 1 and keeping the same width-height ratio, we crop images, which are then resized to the input image size with bi-cubic interpolation. For a given pair of images, we apply identical transformations to both images. 

\begin{figure*}[t]
  \centering
   \includegraphics[width=\linewidth]{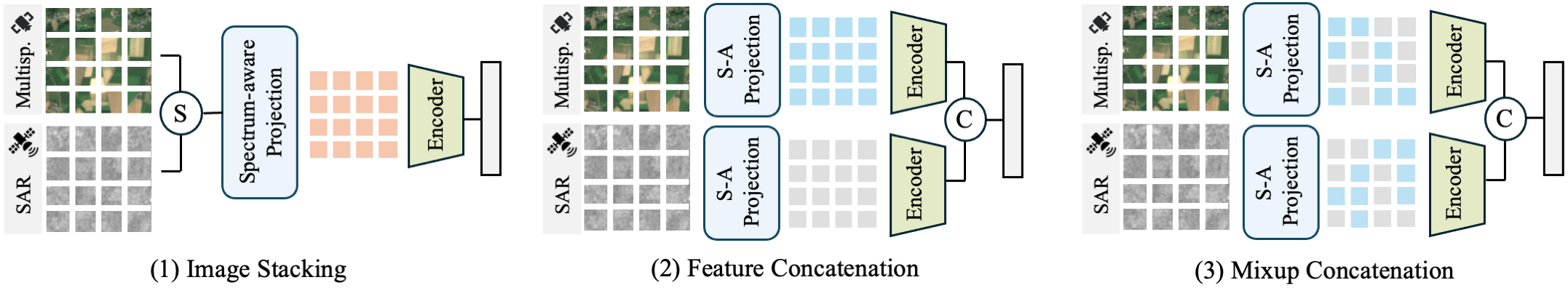}
   \caption{Different multi-modal fusion strategies for downstream transfer on multi-modal input images. S-A Projection, S and C denote Spectrum-aware Projection, stacking and concatenation, respectively.}
   \label{fig:pipeline_inference}
\end{figure*}
\subsection{Evaluation}
\label{supp:eval_sect}
Once SMARTIES is pretrained with ViT-B and ViT-L backbones, the resulting encoders and spectrum-aware projection layers are used for single/multi-modal downstream transfer of single/multi-label classification and semantic segmentation with RS images from diverse sensors. For downstream transfer, we consider all the possible inference modes: (1) \textit{in-domain, seen sensor inference}, (2) \textit{open-domain, seen sensor inference}, and (3) \textit{open-domain, unseen sensor inference} that are illustrated in \cref{fig:setups}. 

For a fair comparison with other foundation models, we follow the same evaluation protocols and the splits of datasets with previous works by using non-parametric \textit{k}NN classification, linear probing, non-linear frozen backbone finetuning and full finetuning. \textit{k}NN classification allows to directly assess the learned representations without additional training, while linear probing or frozen backbone finetuning require to train a linear classifier or nonlinear task head, respectively, on top of the frozen backbone. Finetuning requires to train the entire backbone with the task head on the downstream dataset. Below, we provide the evaluation details for each dataset.

\vspace{.1cm}\noindent\textbf{BigEarthNet-S1.} By following CROMA~\cite{croma}, we apply linear probing by using the 10\% of the complete training set and evaluating on the entire validation set without data augmentation. Linear probing is applied for 100 epochs by using AdamW optimizer with the batch size of 1024 and the base learning rate of 1e-3, which is decayed 10$\times$ at epochs 60 and 80. We resize images to the input image size with bi-cubic interpolation.

\vspace{.1cm}\noindent\textbf{BigEarthNet-S2.} By following SatMAE (S2)~\cite{satmae} and SeCO~\cite{seco}, we apply full finetuning by using the 10\% of the complete training set and evaluating on the entire validation set. We finetune for 100 epochs, using AdamW optimizer with the batch size of 256, the base learning rate of 5e-5, the weight decay of 0.05, the drop path rate of 0.2, warmup of 5 epochs and cooldown by half-cosine decay schedule. For data augmentation, we first randomly apply vertical flipping, horizontal flipping and rotation in order. Then, we resize images to the input image size with bi-cubic interpolation.

\vspace{.1cm}\noindent\textbf{BigEarthNet-MM.} By following CROMA~\cite{croma}, we apply linear probing by using 10\% of the complete training set and evaluating on the entire validation set without data augmentation. Linear probing is applied for 100 epochs by using AdamW optimizer with the batch size of 1024 and the base learning rate of 1e-3, which is decayed 10$\times$ at epochs 60 and 80. We resize image pairs to the input image size with bi-cubic interpolation. To operate SMARTIES on multi-modal input, as shown in~\cref{fig:pipeline_inference}, we consider three multi-modal fusion strategies: 1) image stacking; 2) feature concatenation; and 3) mixup concatenation. Compared to first two strategies, mixup concatenation, where we concatenate features extracted from the backbone from both modalities after applying mixup with spectrum-aware projections, is introduced for the first time in our paper.

\vspace{.1cm}\noindent\textbf{EuroSAT.} We apply linear probing, \textit{k}NN classification and fine-tuning on EuroSAT by using the same splits as SatMAE (S2)~\cite{satmae}. For linear probing, we use AdamW optimizer for 100 epochs with the batch size of 1024 and the base learning rate of 1e-3, which is decayed 10$\times$ at epochs 60 and 80. For finetuning, we use AdamW optimizer for 150 epochs with the batch size of 256, the weight decay of 0.05, the base learning rate of 2e-4, the drop path rate of 0.1, warmup of 5 epochs and cooldown by half-cosine decay schedule. We also apply CutMix ($\alpha=1$) and MixUp ($\alpha=0.8$) with between images and labels. Only for finetuning, we use data augmentation with random vertical flipping, horizontal flipping and rotation in order. We resize images to the input image size with bi-cubic interpolation.

\vspace{.1cm}\noindent\textbf{RESISC-45.} We apply fine-tuning on RESISC-45 by using the same splits as Scale-MAE~\cite{scale_mae}. To this end, we use AdamW optimizer for 200 epochs with the batch size of 64, the weight decay of 0.05, the base learning rate of 6.25e-5, the drop path rate of 0.2, warmup of 5 epochs and cooldown by half-cosine decay schedule. For data augmentation, we first apply random vertical flipping, horizontal flipping and rotation in order. Then, by randomly sampling scale parameter between 0.25 and 1 and keeping the same width-height ratio, we crop images, which are then resized to the input image size with bi-cubic interpolation during training. During evaluation, we first resize images to 256$\times$256, and then apply center cropping with the input image size. 

\begin{table*}
  \centering
  \renewcommand{\arraystretch}{0.92}
  \begin{tabular}{@{}lcccccc@{}}
    \toprule
    \multirow{2}{*}{Model} & \multirow{2}{*}{Backbone} & \multirow{2}{*}{PT Epochs} & \multicolumn{2}{c}{PT Data Size} & mAP (\%) & Acc. (\%)\\
     &  &  & S2 & RGB & BEN-S2 10\% & RESISC-45 \\
    \midrule
    SatMAE (S2)~\cite{satmae} & ViT-L & 50 & 713K & - & 82.1 & \nap \\
    SatMAE (S2)~\cite{satmae} & ViT-L & 200 & 713K & - & 86.2 & \nap\\
    SatMAE (RGB)~\cite{satmae} & ViT-L & 800 & - & 364K & \nap & 94.8\\
    Scale-MAE~\cite{scale_mae} & ViT-L & 800 & - & 364K & \nap & 95.7 \\
    CROMA~\cite{croma} & ViT-B ($\times$2) & 300 & 1M & - & 87.6 & \nap \\
    SpectralGPT~\cite{spectral_gpt} & ViT-L  & 200 & 713K & - & 86.9 & \nap \\
    SpectralGPT$^+$~\cite{spectral_gpt} & ViT-L & 300 & 1M & - & 89.0 & \nap \\
    S2MAE~\cite{s2mae} & ViT-L & 200 & 713K & - & 86.5 & \nap \\
    S2MAE$^*$~\cite{s2mae} & ViT-L & 300 & 1M & - & 88.5 & \nap \\
    SatMAE++ (RGB)~\cite{satmae_pp} & ViT-L & 800 & - & 364K & \nap & 97.5 \\
    SatMAE++ (S2)~\cite{satmae_pp} & ViT-L & 50 & 713K & - & 85.1 & \nap \\
    SMARTIES (Ours) & ViT-L & 300 & 248K & 60K & 87.7 & 95.8 \\ \midrule
    CROMA~\cite{croma} & ViT-L ($\times$2) & 600  & 1M & - & 88.3 & \nap \\
    SpectralGPT~\cite{spectral_gpt} & ViT-H & 200 & 713K & - & 89.2 & \nap \\
    SpectralGPT$^+$~\cite{spectral_gpt} & ViT-H & 300 & 1M & - & 91.4 & \nap \\
    S2MAE~\cite{s2mae} & ViT-H & 200 & 713K & - & 88.8 & \nap \\
    S2MAE$^*$~\cite{s2mae} & ViT-H & 300 & 1M & - & 90.7 & \nap \\
    SkySense~\cite{skysense} & ViT-L ($\times$2) + Swin-H & 780 & 21.5M & 21.5M & 88.7 & 96.3*\\
    \bottomrule
  \end{tabular}
  \caption{Pretraining efficiency comparison of the existing foundation models, including 1) the considered backbones, 2) pretraining (PT) epochs, 3) PT data size in terms of numbers of Sentinel-2 (S2) and RGB images, 4) BEN multi-label scene classification results (mAP) when finetuning (FT) is applied with 10\% of the training set, 5) RESISC-45 scene classification results (top-1 accuracy) under FT. *20\% of the training set is used. {\nap} indicates \textit{not applicable} due to either the lack of publicly available models or sensor mismatch between models and datasets (which could lead to unfair comparisons).}
  \label{tab:efficiency_compare}
\end{table*}

\vspace{.1cm}\noindent\textbf{WHU-RS19.} For \textit{k}NN classification, we only resize images to the input image size with bi-linear interpolation. 

\vspace{.1cm}\noindent\textbf{UCMerced.} We first resize images to 256$\times$256, and then apply center cropping with the input image size for \textit{k}NN classification. 

\vspace{.1cm}\noindent\textbf{BurnScars, DynamicEarthNet, SpaceNet7} experiments are conducted by following the default evaluation protocol of the PANGAEA~\cite{pangaea} benchmark for a fair comparison with other methods. In detail, for all these datasets, frozen backbone UPerNet probing is applied: model weights of our pretrained encoder are frozen, while UPerNet segmentation head is learned on top it. As DynamicEarthNet includes multi-temporal images, Lightweight Temporal Attention Encoder (L-TAE)~\cite{ltae} is utilized between the encoder and the segmentation head to map each image time-series into an aggregated feature map. To learn the segmentation head parameters for all the datasets, AdamW optimizer is used for 80 epochs with the batch size of 8, the weight decay of 0.05 and the base learning rate of 1e-4, which is decayed 10$\times$ at 60\% and 90\% of the total steps. We refer readers to~\cite{pangaea} for the details of the PANGAEA evaluation protocol.

\vspace{.1cm}\noindent\textbf{SICKLE.} We apply non-linear frozen backbone finetuning by freezing the parameters of our pretrained model, while learning a segmentation head on top it. We use the same segmentation head with~\cite{sickle_wacv_2024} by using a single convolutional layer followed by bi-linear upsampling. For zero-shot sensor transfer to Landsat-8 images, we apply interpolation to unseen spectrum ranges of: 1) blue band (B2) via the weighted average of the projection layers dedicated to Sentinel-2 blue (B02) and aerosol (B01) bands; and 2) thermal infrared band (B10) via the weighted average of the projection layers dedicated to Sentinel-2 SWIR band (B12) and Sentinel-1 VV band. For the rest of the bands, we select the relevant projection layers dedicated to Sentinel-2 bands, where the same spectral ranges are shared with Landsat-8 bands. To learn the segmentation head, we use AdamW optimizer for 200 epochs with the batch size of 32 and the base learning rate of 8e-3, which is decayed 10$\times$ at epochs 120 and 160. Without any data augmentation, we resize images to the input image size with nearest-neighbor interpolation. 

\section{Pretraining Efficiency}
\label{supp:pt_efficiency_sec}
In the main body of our paper, we test our models against the existing foundation models, which use as similar pretraining (PT) data size and epochs as possible, for a fair comparison. To further compare the PT efficiency of SMARTIES, in~\cref{tab:efficiency_compare} we provide an extended comparison of the existing models in terms of PT data size and epochs together with BEN-S2 and RESISC-45 results under finetuning. Results demonstrate that SMARTIES shows a significantly higher PT efficiency compared to previous methods in terms of both PT data size and epochs (which is associated with PT time). In detail, by comparing the results in the first block of~\cref{tab:efficiency_compare}, one can see that SMARTIES uses the fewest Sentinel-2 (S2) images for PT (248K) to achieve highly competitive performance (87.7\%) on BEN-S2 compared to the state-of-the-art SpectralGPT$^+$ model (89.0\%), which uses four times more of S2 images during PT. Meanwhile, SMARTIES also shows high efficiency in terms of the use of RGB data. By using only 60K RGB PT data, SMARTIES surpasses most of the RGB-specific models pretrained with 6 times more data and over 2 times more PT epochs. The high data efficiency of SMARTIES can be attributed to: 1) the sensor-agnostic design, which explicitly represents data into transferable spectrum-aware spaces instead of learning shared representations from heterogeneous sensors implicitly; and 2) the implicit data augmentation brought by cross-sensor token mixup. We would like to note that masked data modeling in combination with ViTs can be effectively scaled into larger models with higher amount of PT data~\cite{mae_2022_cvpr}. This can be seen from~\cref{tab:efficiency_compare}: 50 vs. 200 epochs PT of SatMAE (S2) and S2MAE (ViT-L) vs. S2MAE (ViT-H). Thus, by feeding more PT data with more epochs, the performance of SMARTIES can be further scaled up. To further analyze this, we assess the effect of different subsets of our PT data together with different PT epochs and backbones in~\cref{tab:ablation_eff} under \textit{k}NN classification of EuroSAT. One can observe from the table that the higher the number of images and epochs used by SMARTIES for PT the better it performs. In addition, by using a larger ViT model, SMARTIES is capable of achiving higher \textit{k}NN accuracy.

\begin{table}
  \centering
  \renewcommand{\arraystretch}{0.9}
  \setlength\tabcolsep{8pt}
  \begin{tabular}{@{}lcccc@{}}
    \toprule
    \multirow{2}{*}{Backbone} & \multirow{2}{*}{PT Epochs} & \multicolumn{2}{c}{PT Data Split} & \multirow{2}{*}{Acc.} \\
    & & BEN & fMoW & \\
    \midrule
    \multirow{4}{*}{ViT-B} & \multirow{3}{*}{50} & \cmark & \xmark & 91.1\\
    & & \xmark & \cmark & 92.1 \\
    & & \cmark & \cmark & 93.2\\\cmidrule{2-5}
    & 100 & \cmark & \cmark & 94.3\\\midrule
    ViT-L & 100 & \cmark & \cmark & 94.6\\
    \bottomrule
  \end{tabular}
  \caption{\textit{k}NN classification accuracy (\%) on EuroSAT when different subsets of the pretraining (PT) data are used for SMARTIES.}
  \label{tab:ablation_eff}
\end{table}

\begin{table}
  \centering
  \setlength\tabcolsep{1pt}
  \renewcommand{\arraystretch}{0.94}
  \begin{tabular}{@{}lcccc@{}}
    \toprule
    Method & Backbone & PT Epochs & SAR PT & mAP\\
    \midrule
    SMARTIES (w/o PE) & ViT-B & 50 & \xmark & 62.1\\
    SMARTIES (w PE) & ViT-B & 50 & \xmark & 64.0 \\
    SMARTIES (Ours) & ViT-B & 50 & \cmark & 73.6 \\ 
    \midrule
    SpectralGPT~\cite{spectral_gpt} & ViT-B & 200 & \xmark & 57.1\\
    SatMAE (S2)~\cite{satmae} & ViT-L & 200 & \xmark & 67.4 \\
    SMARTIES (Ours) & ViT-B & 300 & \cmark & 78.9 \\
    \bottomrule
  \end{tabular}
  \caption{BEN-S1 multi-label classification results (mAP) when linear probing is applied with 10\% of the training set. PE: projection extrapolation; PT: pretraining.}
  \label{tab:ben_s1}
\end{table}

\section{Extrapolation to Unseen Spectral Ranges}
\label{supp:extrapolation_sect}
For downstream transfer to a sensor unseen during pretraining (i.e., open-domain, unseen sensor inference), SMARTIES can be adapted to unseen spectral ranges by applying interpolation to the learned projection layers as it is explained in Sec. 3.4 and shown with the SICKLE results (cf. Sec. 4.4). Here, we further evaluate the generalization ability of SMARTIES to unseen regions out of the (min, max) of the pretraining spectra through extrapolation. This setting is significantly more challenging than the SICKLE experiments, where the thermal infrared bands falls within the pretraining range. We simulate an unseen spectral range out of the pretraining spectra by excluding SAR data during pretraining. Then, we perform linear probing on BEN-S1 by extrapolating the learned projection layers of SMARTIES. In addition, we also apply linear probing with SpectralGPT and SatMAE (S2), for which SAR is already excluded from pretraining. To do this, we duplicate VV and VH bands of BEN-S1 six times, which are given as model inputs in place of the original input (Sentinel-2 image). Table~\ref{tab:ben_s1} shows the corresponding results. One can see from the table that SMARTIES pretrained without SAR data yields lower BEN-S1 results than the full pretraining even though projection extrapolation yields modest +2\% mAP. This shows that the downstream transfer capability of SMARTIES to an unseen sensor is valid: 1) for the the unseen ranges falling inside the pretraining spectra through projection interpolation; 2) not for the ranges out of the limits of pretraining spectra through projection extrapolation. Once SAR data is included in pretraining, however, SMARTIES with even 50 pretraining epochs provides 16.5\% higher mAP than SpectralGPT, which rely on sensor-specific pretraining. These results indicate that the generalization capability of SMARTIES highly depends on the spectral range seen during pretraining.



\end{document}